\documentclass{article}

\PassOptionsToPackage{numbers, compress}{natbib}
\usepackage[preprint]{neurips_2026}

\usepackage[utf8]{inputenc} 
\usepackage[T1]{fontenc}    
\usepackage{hyperref}       
\usepackage{url}            
\usepackage{booktabs}       
\usepackage{amsfonts}       
\usepackage{nicefrac}       
\usepackage{microtype}      
\usepackage{xcolor}         

\usepackage{amsmath}
\usepackage{amssymb}
\usepackage{mathtools}
\usepackage{amsthm}

\usepackage[capitalize,noabbrev]{cleveref}

\theoremstyle{plain}
\newtheorem{theorem}{Theorem}[section]

\theoremstyle{definition}

\theoremstyle{remark}

\usepackage[textsize=tiny]{todonotes}

\usepackage{enumitem}
\usepackage{CJKutf8}
\usepackage{makecell}
\usepackage{tabularx}
\usepackage{colortbl}
\usepackage{subfigure}
\usepackage{multirow}
\usepackage{wrapfig}
\usepackage{tcolorbox}
\tcbuselibrary{breakable,skins}
\usepackage{arydshln}
\usepackage{bbding}
\usepackage{fontawesome5} 

\usepackage[table]{xcolor} 
\usepackage{ulem} 
\usepackage{tikz} 

\definecolor{mlb}{RGB}{173,216,230}  
\definecolor{mlo}{RGB}{255,223,186}  
\definecolor{varblue}{RGB}{0, 120, 215}

\newtcolorbox{promptbox}[1]{
    title=#1,
    colback=blue!5!white,       
    colframe=blue!75!black,     
    coltitle=white,             
    breakable,
    fontupper=\small
}

\title{Agentic Reward Modeling: Verifying GUI Agent via Progressive Trajectory-Grounded Interaction}

\author{
\parbox{\textwidth}{
\raggedright
Chaoqun Cui$^{1,2,*}$,~Jing Huang$^{3,*}$,~Shijing Wang$^{4}$,~Liming Zheng$^{3}$,~Qingchao Kong$^{1,2}$, \\
~Zhixiong Zeng$^{3,\dagger}$ \\
\normalfont\mdseries
$^{1}$MAIS, Institute of Automation, Chinese Academy of Sciences \\
$^{2}$School of Artificial Intelligence, University of Chinese Academy of Sciences \\
$^{3}$Meituan \\
$^{4}$School of Computer Science \& Technology, Beijing Jiaotong University \\
\texttt{cuichaoqun2025@ia.ac.cn, zengzhixiong@meituan.com} \\
{\footnotesize
$^{*}$ Equal contribution. \quad
$^{\dagger}$ Corresponding author.
}\\[2pt]
}
}

\begin{document}

\maketitle

\begin{abstract}

Reinforcement learning with verifiable rewards (RLVR) provides a promising pathway for continuously advancing GUI agents, yet existing reward modeling paradigms face complementary limitations. Rule-based methods suffer from poor scalability and cannot handle open-ended tasks. LLM-as-a-Judge methods enable scalable trajectory verification but remain passive and are constrained by partial state observability, since key evidence often resides in latent environment states beyond the trajectory. Recent active environment interaction methods mitigate observability issues but tend to over-rely on probing while under-utilizing direct trajectory evidence, leading to verification inefficiency. To address these challenges, we advocate a trajectory-grounded interactive verification paradigm. We introduce VAGEN, a framework that employs a tool-augmented verifier agent governed by a Progressive Verification Mechanism, which follows a surface-to-latent and cheap-to-expensive design philosophy to extract trajectory evidence and probe environment states in a proactive end-to-end manner. Experiments on OSWorld-Verified and AndroidWorld benchmarks demonstrate that VAGEN significantly improves evaluation accuracy with a favorable performance-efficiency trade-off.

\end{abstract}

\section{Introduction}

Reinforcement learning with verifiable rewards (RLVR) has emerged as a promising framework for the continual improvement of graphical user interface (GUI) agents \cite{webrl,rlvr3,magicgui}. In this setting, a Large Language Model (LLM)-based agent serves as the policy network, directly operating a visualized digital interactive environment (e.g., Windows, Linux, Android) to complete user-specified tasks \cite{uitars2,opencua}. A reliable binary reward signal, indicating whether an execution trajectory successfully completes the task, is therefore essential for providing supervision that promotes successful behaviors while suppressing incorrect ones \cite{vr1,vr2,vr3,vr4}. Existing reward modeling methods are typically categorized into rule-based and LLM-as-a-Judge approaches \cite{prore,smartsnap}, as shown in Figure~\ref{fig:f1} and \ref{fig:f2}.

\textit{Rule-based verification} relies on human experts to manually write task-specific checkers that verify whether the expected state of each task has been achieved. For instance, in OSWorld-Verified \cite{osworld_verified} and AndroidWorld \cite{androidworld} benchmarks, each task is associated with corresponding manually engineered unit testing code/scripts to verify completion. This approach primarily suffers from two major limitations: 1) it fails to provide accurate assessments for open-ended tasks or tasks with multiple valid solutions; and 2) it necessitates substantial human effort, thereby severely restricting its scalability and rendering it inapplicable to large-scale agentic RL training processes.

To achieve scalable rewards, many \textit{LLM-as-a-Judge based trajectory verification} methods are proposed \cite{llmjudge1,digirl,cuarewardbench,osthemis}. These methods leverage the semantic and visual understanding capabilities of models such as GPT-4o \cite{gpt4o} to observe and seek evidence of whether the task is completed from the reasoning text and screenshots of task trajectories. Although this paradigm is scalable and efficient, enabling the extraction of direct evidence from trajectories, it still faces two fundamental limitations: 1) It is inherently \textit{passive}. The judge model can only consume the pre-recorded trajectory, which often consists of dozens of interaction steps with both visual and textual information, but cannot autonomously decide which intermediate states should be revisited or examined in greater detail. As a result, critical evidence may be diluted in long-horizon trajectories, while irrelevant or noisy steps may distract the judge. 2) Such methods suffer from \textit{partial state observability} of agent trajectories. In certain tasks, the visual observation of the trajectory can only reflect the \textit{surface state} of the GUI environment, while the key evidence of task completion is hidden in the \textit{latent environment states}, such as file-system changes, application configurations, or background processes. In other words, LLM-as-a-Judge often relies on explicit visual evidence, whereas for certain tasks, the decisive indicators of success may lie in implicit system state changes, and even human evaluators find it difficult to make judgments based solely on these surface visual feedbacks.

To overcome the limitations of passive trajectory verification, recent studies have introduced \textit{active environment interaction} into GUI reward modeling \cite{prore,smartsnap}. This paradigm employs a dedicated GUI agent as the verifier, allowing it to interact directly with the environment to inspect latent environment states that are invisible in the recorded trajectory, thereby alleviating the partial state observability issue. However, this active probing paradigm also introduces new challenges: 1) Since verification requires extra agent execution in the environment, it can be substantially \textit{less efficient} than direct trajectory-based judgment, especially for multi-state tasks (e.g., checking multiple success conditions) or complex environments (e.g., desktop environments such as OSWorld) that require several rounds of probing. 2) Moreover, existing interaction-heavy methods may \textit{over-rely on environment probing} while under-utilizing the direct evidence already present in the trajectory, such as explicit intermediate observations or final visual states. As a result, the verifier may perform redundant or weakly grounded probing even when the trajectory itself contains sufficient clues for reliable judgment. 

\begin{wraptable}{r}{0.45\textwidth}
\vspace{-1.8em}
\centering
\caption{Validation of the "easy to verify, hard to solve" property. Experiments using the Claude-Sonnet-4.5 model.}
\resizebox{0.45\textwidth}{!}{
\begin{tabular}{lcc|cc}
\Xhline{1.0pt}
\rowcolor{gray!20}
~ & \multicolumn{2}{c}{\textbf{OSWorld-Verified}} & \multicolumn{2}{c}{\textbf{AndroidWorld}} \\
\cline{2-5} 
\rowcolor{gray!20}
\multirow{-2}{*}{\textbf{Role}} & \textbf{SR (\%)} & \textbf{Avg Steps} & \textbf{SR (\%)} & \textbf{Avg Steps} \\
\hline
Actor & 55.9 & 28.5 & 62.1 & 23.6 \\
Verifier & 83.1 & 17.4 & 93.1 & 16.5 \\
\Xhline{1.0pt}
\end{tabular}
}
\vspace{-0.5em}
\label{tab:easyveri}
\end{wraptable}

In light of the above observations, we advocate a \textit{trajectory-grounded interactive verification} paradigm that unifies trajectory verification and environment interaction. This paradigm can efficiently and selectively extract direct evidence from trajectories, while probing latent system states through environment interaction to mitigate partial state observability in trajectories. Specifically, we propose \textbf{V}erification via \textbf{A}gentic Trajectory-\textbf{G}rounded \textbf{E}nvironment I\textbf{n}teraction (VAGEN) framework. We instantiate a tool-augmented verifier agent equipped with tools for trajectory inspection and GUI environment interaction. The verification process follows a Progressive Verification Mechanism, which adopts a surface-to-latent and cheap-to-expensive design philosophy. The verifier agent performs verification in a proactive end-to-end manner: it first autonomously inspects the trajectory and early-stops once sufficient evidence is found to avoid additional cost. When trajectory evidence is insufficient, it conducts interactive verification by executing shell or Python commands or directly manipulating the GUI environment. The feasibility of this agent-based verification is supported by a property observed in many tasks, namely "easy to verify, hard to solve" \cite{easyveri1,easyveri2,easyveri3}. For example, for the task "Help me buy the book Reinforcement Learning by Richard.", the actor agent needs to complete many steps, such as opening an online shopping app, searching, placing an order, and making a payment, whereas the verifier agent only needs to check whether the corresponding item has been purchased on the order page. This indicates that the verifier agent can complete the verification process efficiently. Table~\ref{tab:easyveri} shows that the verifier agent achieves a higher success rate (SR) than the actor agent, thereby validating this property.

In summary, the contributions of this study are as follows:
\begin{itemize}
\item We identify the complementary strengths and limitations of passive trajectory verification and active environment interaction, and advocate a trajectory-grounded interactive verification paradigm that unifies both for reliable GUI agent reward modeling.
\item We propose VAGEN, a framework that instantiates this paradigm via a tool-augmented verifier agent governed by a Progressive Verification Mechanism, which follows a surface-to-latent and cheap-to-expensive design philosophy to achieve end-to-end verification.
\item Extensive experiments demonstrate that VAGEN consistently outperforms both passive and active verification baselines while achieving a favorable performance-efficiency trade-off, and further boosts performance via a test-time scaling strategy.
\end{itemize}

\section{Related Work}

This section reviews recent advances in GUI agents and verifiable reward modeling, with a focus on trajectory-based and interaction-based verification methods, and clarifies the positioning of VAGEN.

\subsection{Computer Use Agent for GUI Automation Tasks}

The capabilities of GUI agents are built upon the visual understanding and reasoning abilities of Vision-Language Models (VLMs). They typically follow the ReAct paradigm \cite{react}, in which they observe the GUI screenshot, reason via chain of thought, and act at each step of an execution trajectory. Recently, research related to GUI agents advances rapidly. Research works such as the UI-TARS series \cite{uitars,uitars2}, OpenCUA \cite{opencua}, UITron \cite{uitron}, Step-GUI \cite{stepgui}, and MAI-UI \cite{maiui} systematically investigate data construction, cleaning pipelines, and training methods covering both pre-training and post-training stages. Meanwhile, mainstream foundation models like the Claude series \cite{claude}, Seed series \cite{seed}, and Qwen series \cite{qwen25,qwen3} begin to natively integrate GUI interaction capabilities. Furthermore, works utilizing multi-agent collaboration and hierarchical architectures provide valuable insights for the research community \cite{agents,agents2,agents3,mav3}.

\subsection{Verifiable Rewards for GUI Agents}

Verifiable rewards are crucial for the continuous evolution of GUI agents \cite{uitars2,vr4}. Early methods primarily rely on rule-based verification \cite{osworld,androidworld,windows}. Although these methods possess high accuracy, they are difficult to scale to diverse open-ended tasks \cite{cuarewardbench}. To obtain scalable rewards, further studies utilize VLMs as reward models \cite{survey}, including Outcome Reward Models (ORM) for offline trajectory filtering, such as DigiRL \cite{digirl}, WebRL \cite{webrl}, and ZeroGUI \cite{zerogui}, and Process Reward Models (PRM) that provide step-wise signals for online reinforcement learning, such as SEAgent \cite{seagent}, GUI-OWL \cite{mav3}, Orcust \cite{orcust}, GUI-PRA \cite{guipra}, and ADMIRE \cite{admire}. However, these VLM-based methods mainly perform passive trajectory verification and face partial state observability issue. To mitigate this issue, recent studies introduce active environment interaction into reward modeling. ProRe \cite{prore} adopts a reasoner-actor architecture in which a general-purpose reasoner schedules state-probing tasks and a domain-specific evaluator agent executes them, judging task success by checking the consistency between trajectory-derived claims and probing-derived claims. Nevertheless, its verification process largely relies on environment probing, while trajectory is mainly abstracted into claims for consistency checking with probing results, which may under-utilize direct evidence already present in trajectories. SmartSnap \cite{smartsnap} couples task execution and verification within a learned policy through joint training, which constrains its practical applicability and makes it unsuitable as a plug-in verifier baseline in our experiments. These limitations motivate a trajectory-grounded interactive verification paradigm that jointly exploits trajectory evidence and environment interaction.

\section{Method}

This section introduces the formulation, framework, key components, and scaling strategies of VAGEN.

\subsection{Problem Formulation}

\noindent \textbf{Actor Agent Trajectory.} Given task $q$, \textit{actor agent} $\pi _{a}$ interacts with environment $\mathbb{E}$ over multiple steps to complete the task. At the $i$-th step, $\pi _{a}$ generates reasoning $r_i$ based on current state $s_i$ of the environment and outputs an executable action $a_i$. Upon executing $a_i$ in the environment, the state transitions to $s_{i+1}=\mathbb{E}(s_i,a_i)$. This forms a trajectory $\mathcal{T}$:
\begin{equation}
\mathcal{T}=\{q,s_1,(r_1,a_1),\cdots ,s_{n-1},(r_{n-1},a_{n-1}),s_n\}.
\end{equation}
Terminal state $s_n$ usually contains critical evidence for verifying whether a task is completed. Each state $s_i$ includes an RGB screenshot of current interface (interchangeably referred to as $s_i$ hereafter), serving as a crucial medium for observing intermediate states during verification.

\noindent \textbf{Reward Model Formulation.} Given a trajectory $\mathcal{T}$, a reward model $\mathcal{R}$ is used to predict whether the trajectory $\mathcal{T}$ completes the task $q$:
\begin{equation}
\mathcal{R}(\mathcal{T})=\hat{R}\in \{0, 1\}
\end{equation} 
where $0=\text{failure}, 1=\text{success}$. In VAGEN, $\mathcal{R}$ is instantiated as a trajectory-grounded verifier agent $\pi_e$ that can interact with the environment when needed.

\subsection{Overall Framework}

\begin{figure*}[!t]
  \centering
  \subfigure[Rule-based Checking\label{fig:f1}]{\includegraphics[width=0.325\textwidth]{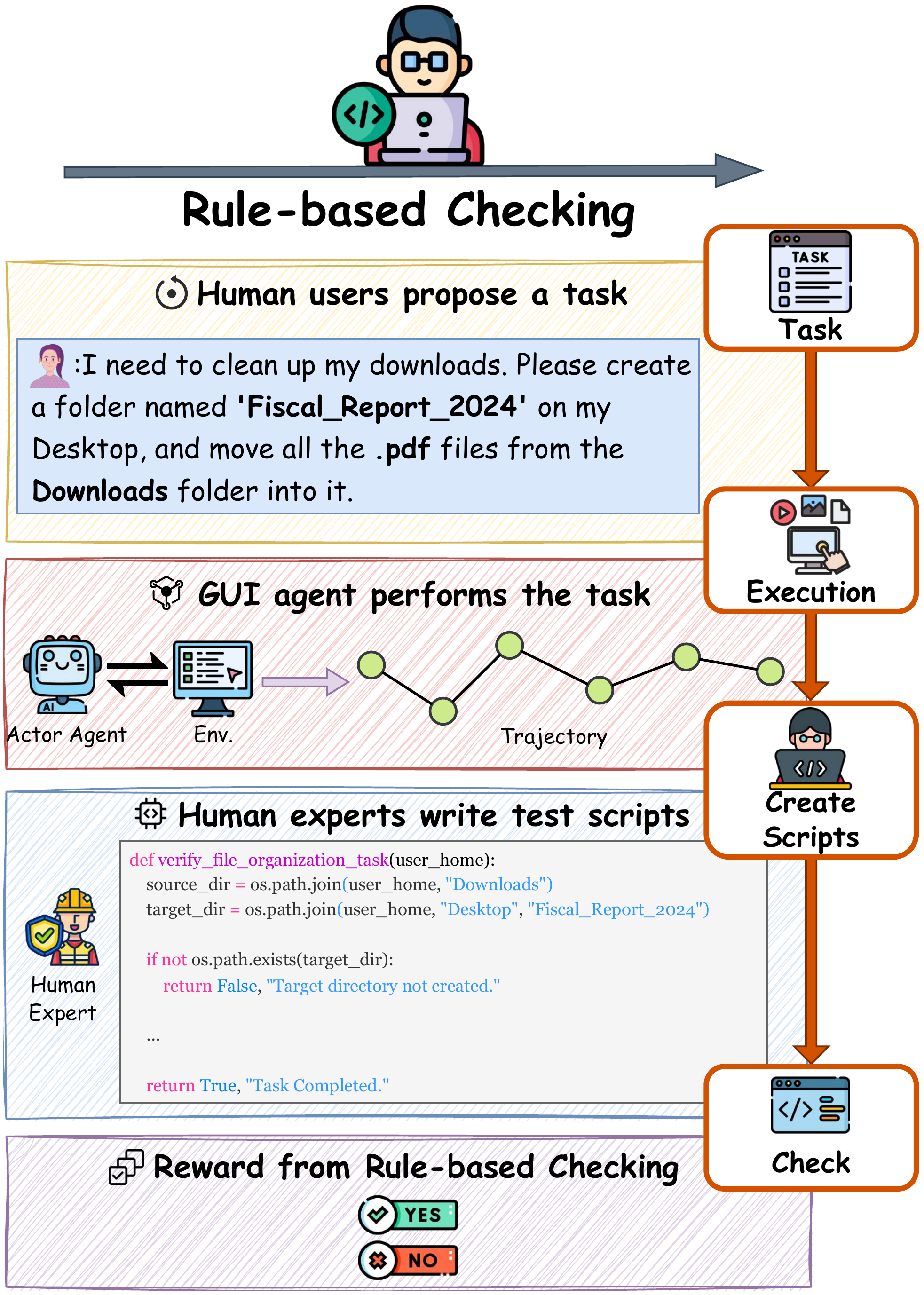}}
  \subfigure[LLM-as-a-Judge\label{fig:f2}]{\includegraphics[width=0.325\textwidth]{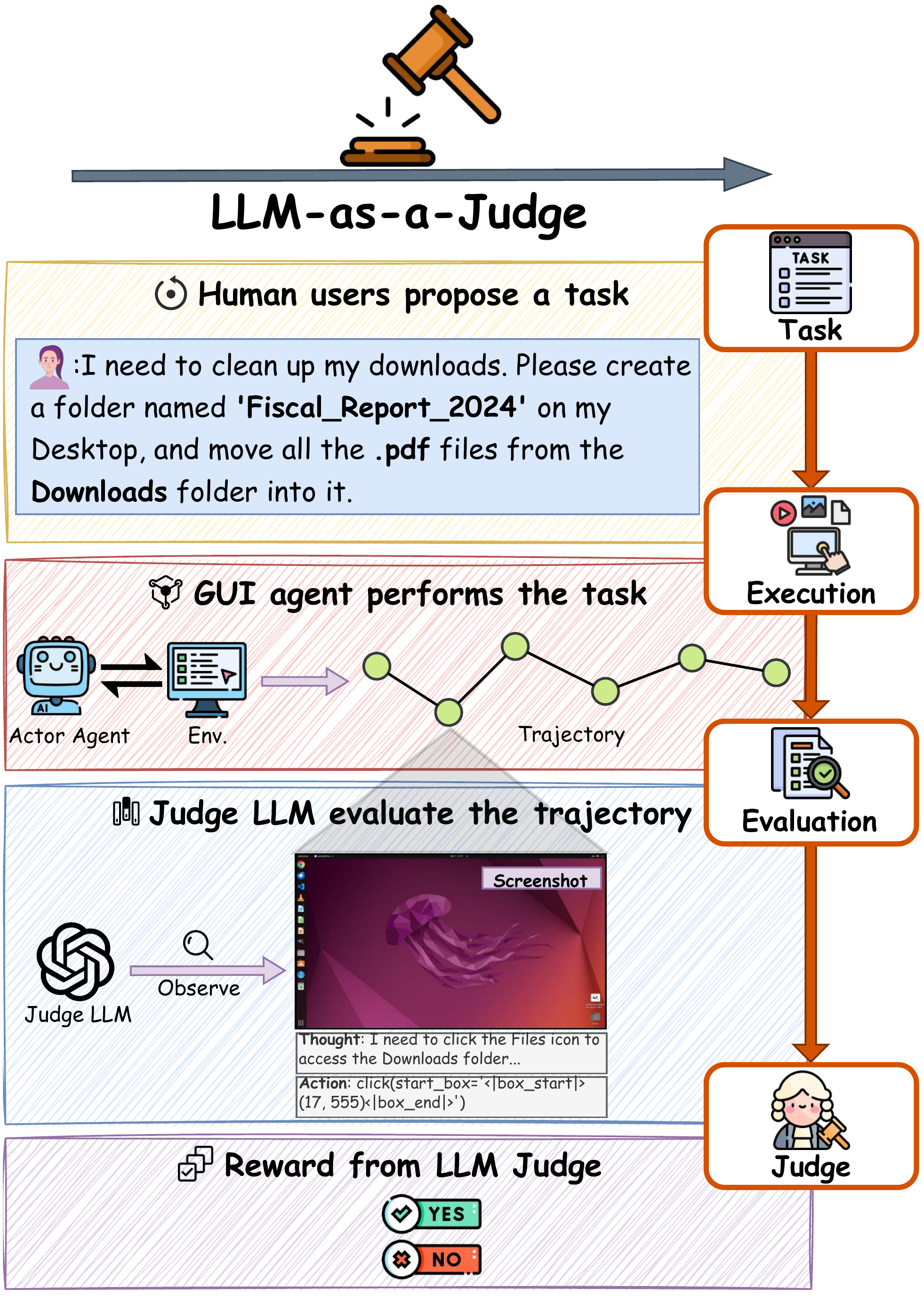}}
  \subfigure[Trajectory-Grounded Interaction\label{fig:f3}]{\includegraphics[width=0.325\textwidth]{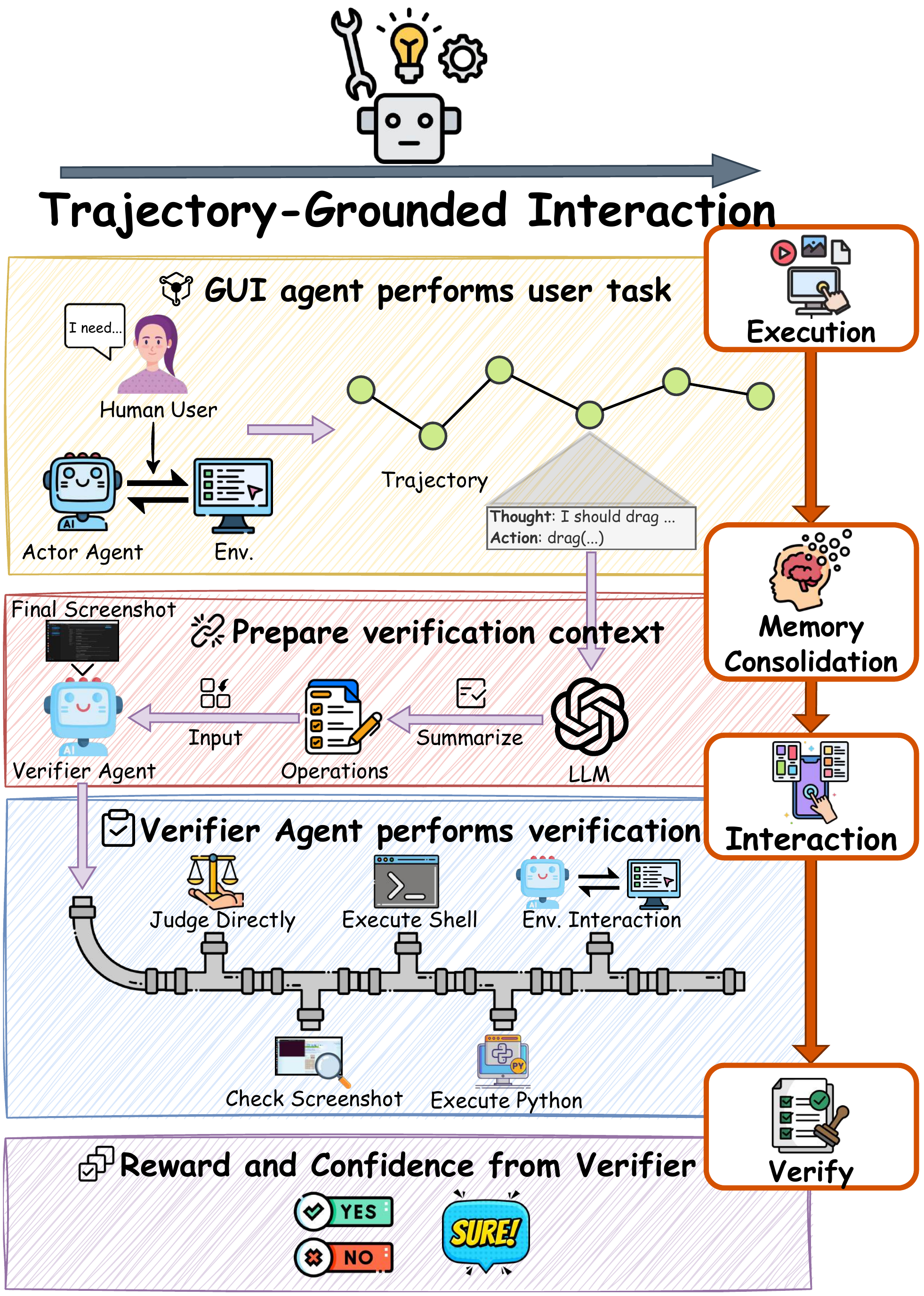}}
  \caption{The evolution of GUI agent verification methods. Unlike (a) Rule-based and (b) LLM-as-a-Judge approaches which suffer from scalability or partial observability issues, (c) Trajectory-Grounded Interactive Verification (VAGEN) instantiates a tool-augmented verifier agent capable of trajectory verification and environment interaction.}
  \label{fig:framework}
\end{figure*}

We present the overall framework of VAGEN in Figure~\ref{fig:f3}. First, we utilize an LLM to summarize the actor agent's trajectory for memory management (Section~\ref{sec:memory}). Subsequently, we employ a tool-augmented agent equipped with various verification tools as the verifier agent (Section~\ref{sec:tool}), which follows a progressive verification mechanism during the verification process (Section~\ref{sec:progressive}). Finally, the verifier agent makes a judgement on the evaluated trajectory, providing a reward and confidence.

\subsection{Trajectory Memory Consolidation}
\label{sec:memory}

Generally, the reasoning $r_i$ at the $i$-th step of the actor agent's trajectory can be divided into three main components (although not necessarily present in every step):
\begin{itemize}
\item \textbf{State Observation} (\textit{What happened on the screen}): the agent's description of the current environment state (screenshot) and the reception of action feedback from the previous step.
\item \textbf{Sub-goal Analysis} (\textit{What the agent plans to do}): the agent's plan and intent for the sub-goals of the current step, including task decomposition, error diagnosis, and path routing, etc.
\item \textbf{Action Description} (\textit{Which atomic action the agent wants to execute}): the agent's description of the atomic action it wants to execute at the current step.
\end{itemize}

We use an LLM $\pi _{s}$ to summarize the \textit{state observation} and \textit{action description} within the reasoning $r_i$ (see the prompt in Appendix~\ref{sec:mcp}), while discarding the \textit{sub-goal analysis}. This allows the verifier agent to focus on the actor agent's objective operations rather than its subjective reasoning content (which may contain erroneous thoughts or noise). The verifier agent plans verification actions under the Progressive Verification Mechanism based on the actor’s operations. The process of summarizing reasoning and action into \textit{operations} $\mathcal{H}=\{o_i\}_{i=1}^n = \pi_{s} \left( \{(r_i, a_i)\}_{i=1}^n \right)$ is shown below:

\begin{tcolorbox}[title=Memory Consolidation]
\textbf{Reasoning}: \textcolor{magenta}{Good! I can see your desktop with a notification about software updates.} \textcolor{blue}{I'll help you install Spotify. The easiest way on Ubuntu is through Snap, which is already available on your system.} \textcolor{green}{Let me open a terminal and install it for you.}\\
\textbf{Action}: hotkey(key="Ctrl+Alt+T")
\tcbline
\textbf{Summary}: \textcolor{magenta}{There is a software update notification on the desktop.} \textcolor{green}{The agent opened a terminal using the "Ctrl+Alt+T" hotkey.}
\end{tcolorbox}

\subsection{Verifier Action Space}
\label{sec:tool}

We instantiate the verifier agent $\pi_{e}$ as a tool-augmented agent capable of both trajectory retrospection and proactive environment interaction. Specifically, we adopt a powerful agent model (such as Claude-Sonnet-4.5) and equip it with the following four essential tools:
\begin{itemize}
\item \textbf{Check Screenshot}: Retrieves screenshots from specific steps in the actor agent's trajectory to visually inspect evidence of task completion. 
\item \textbf{Execute Shell}: Executes shell commands within the terminal to verify latent system states (e.g., file attributes, background processes). 
\item \textbf{Execute Python}: Runs Python code to perform complex logical checks or data processing for judgement. 
\item \textbf{Computer Use}: Enables direct interaction with the environment, sharing the identical action space as actor agent, to proactively probe for verification signals.
\end{itemize}

The \textit{check screenshot} tool is utilized to statically check the trajectory, while the other three tools are employed for dynamic verification involving interaction with the environment. It is worth noting that in mobile environments (e.g., AndroidWorld) where command-line and code execution interfaces are not natively accessible, we exclude the \textit{execute shell} and \textit{execute python} tools.

\subsection{Progressive Verification Mechanism}
\label{sec:progressive}

To balance verification efficiency with rigorous state evaluation, the tool configuration above is designed, and VAGEN employs a Progressive Verification Mechanism. This protocol guides the verifier agent $\pi_{e}$ to seek evidence in a progressive manner, transitioning from surface-level observations to deep latent state inspections. Taking $(q,s_n,\mathcal{H})$ as the initial input, $\pi_{e}$ executes the verification process, ultimately yielding the output $(\hat{R},\hat{C})$. Here, $q$ denotes the user task, $s_n$ represents the terminal screenshot, $\mathcal{H}=\{o_i\}_{i=1}^n$ refers to the consolidated operations, $\hat{R}\in \{0, 1\}$ is the reward, and $\hat{C}\in \{\text{LOW},\text{MEDIUM},\text{HIGH}\}$ is the confidence. The entire process unfolds in the following three stages.

\noindent \textbf{Stage 1: Static Assessment.} This stage is defined as a tool-free decision function $\Phi_{\text{static}}$. The verifier agent first conducts a rapid evaluation based solely on the terminal screenshot and the operation summary. This phase aims to identify immediate successes or obvious failures without invoking external tools. If the verifier agent \textit{directly identifies clear evidence}, it will output the result:
\begin{equation}
(\hat{R}_1, \hat{C}_1)=\Phi_{\text{static}}(q, s_n, \mathcal{H}).
\end{equation}

\noindent \textbf{Stage 2: Visual Retrospection.} If the terminal state is ambiguous or lacks context, the process proceeds to the next stage $\Phi_{\text{retro}}$. The verifier agent invokes the \textit{check screenshot} tool based on $\mathcal{H}$ to inspect historical screenshot $s_t$ ($t<n$) within $\mathcal{T}$, thereby constructing a set of key intermediate evidence $\mathcal{E}_{\text{visual}}$ (serving as \textit{indirect supportive evidence}). At this stage, the agent may also output the result:
\begin{equation}
\begin{aligned}
\mathcal{E}_{\text{visual}} = \{s_t &\mid t \in \text{Indices selected by } \pi_e(\mathcal{H})\}\\
(\hat{R}_2, \hat{C}_2) &= \Phi_{\text{retro}}(q, s_n, \mathcal{H}, \mathcal{E}_{\text{visual}}).
\end{aligned}
\end{equation}

\noindent \textbf{Stage 3: Proactive Probing.} When visual history is insufficient to confirm success (\textit{no direct or indirect evidence}), particularly for tasks involving non-visual system changes, the verifier agent transitions to proactive interaction defined as $\Phi_{\text{probe}}$. In this phase, the agent employs the other tools to actively query the latent state of the environment $\mathbb{E}$ (e.g., verifying file attributes, checking background configurations). This process generates a sequence of interactive verification operations aimed at uncovering deep latent evidence $\mathcal{E}_{\text{latent}}$, upon which the final judgment is based:
\begin{equation}
\begin{aligned}
\mathcal{E}_{\text{latent}} &= \text{Interact}(\mathbb{E}, \mathcal{A}_{\text{probe}}) \\
(\hat{R}_3, \hat{C}_3) = &\Phi_{\text{probe}}(q, s_n, \mathcal{H}, \mathcal{E}_{\text{visual}}, \mathcal{E}_{\text{latent}}),
\end{aligned}
\end{equation}
where $\mathcal{A}_{\text{probe}}=\{\textit{execute shell}, \textit{execute python}, \textit{computer use}\}$ represents the set of proactive interaction tools. The final output $(\hat{R}, \hat{C})$ corresponds to the result from the deepest stage triggered during the process. The complete workflow of Progressive Verification Mechanism is shown in Figure~\ref{fig:progressive}. The prompt of verifier agent is shown in Appendix~\ref{sec:vap}.

\begin{figure*}[!h]
  \centering
  \includegraphics[width=\textwidth]{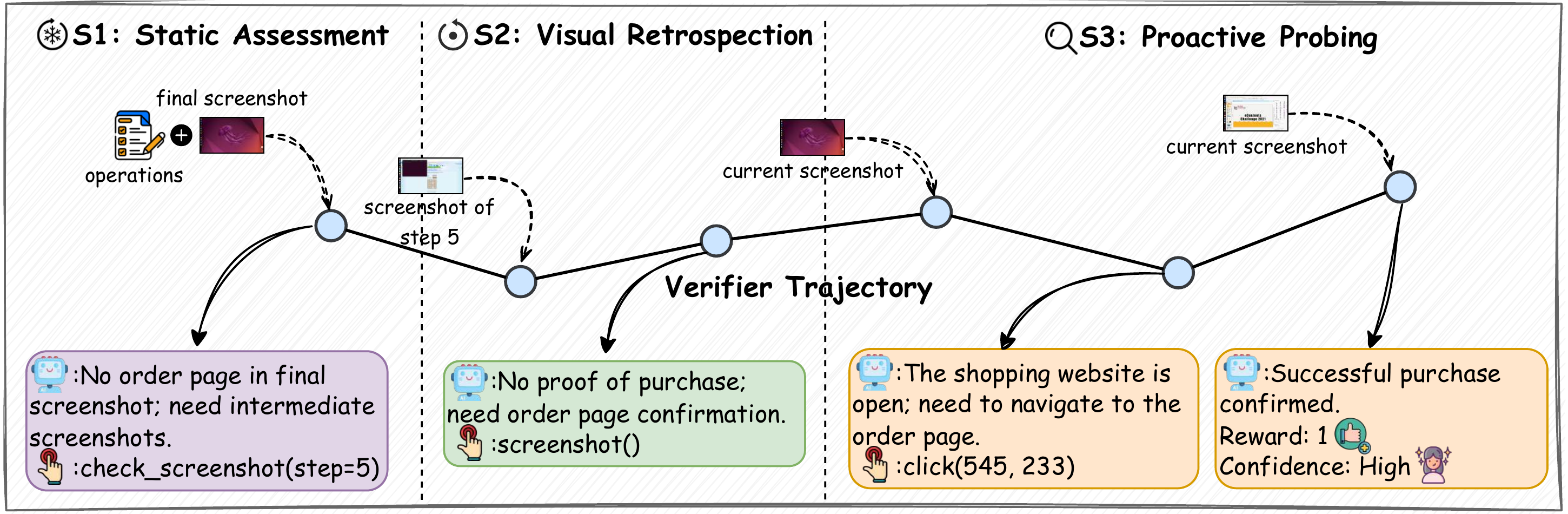}
  \caption{Workflow of the Progressive Verification Mechanism for the task "Help me buy the book Reinforcement Learning by Richard".}
  \label{fig:progressive}
\end{figure*}

\subsection{Test-Time Scaling}

\subsubsection{Read-Only Scaling of Verifier Agent}
\label{sec:readonly}

Directly applying test-time scaling to the verifier agent is challenging due to the high cost of environment state restoration. Restoring the trajectory $\mathcal{T}$'s terminal state via virtual machine snapshots incurs significant I/O latency and storage overhead, while re-executing the trajectory's actions is prone to non-deterministic errors. However, we observe that the verification process is predominantly \textit{inspection-oriented}, involving mostly \textit{read} operations (e.g., checking file and website content) rather than state-altering \textit{write} operations (e.g., creating files, purchasing items). Leveraging this insight, we propose a \textit{Read-Only Scaling} strategy. By restricting the verifier agent to a read-only action space during the scaling phase, we ensure the environment state remains invariant. This allows for the efficient serial execution of multiple verification samples ($N$ times) on a single environment instance without the need for intermediate resets.

\subsubsection{Reward-Guided Scaling of Actor Agent}
\label{sec:rewardguided}

\begin{wrapfigure}{r}{0.48\textwidth}
    \vspace{-4.5em}
    \includegraphics[width=0.48\textwidth]{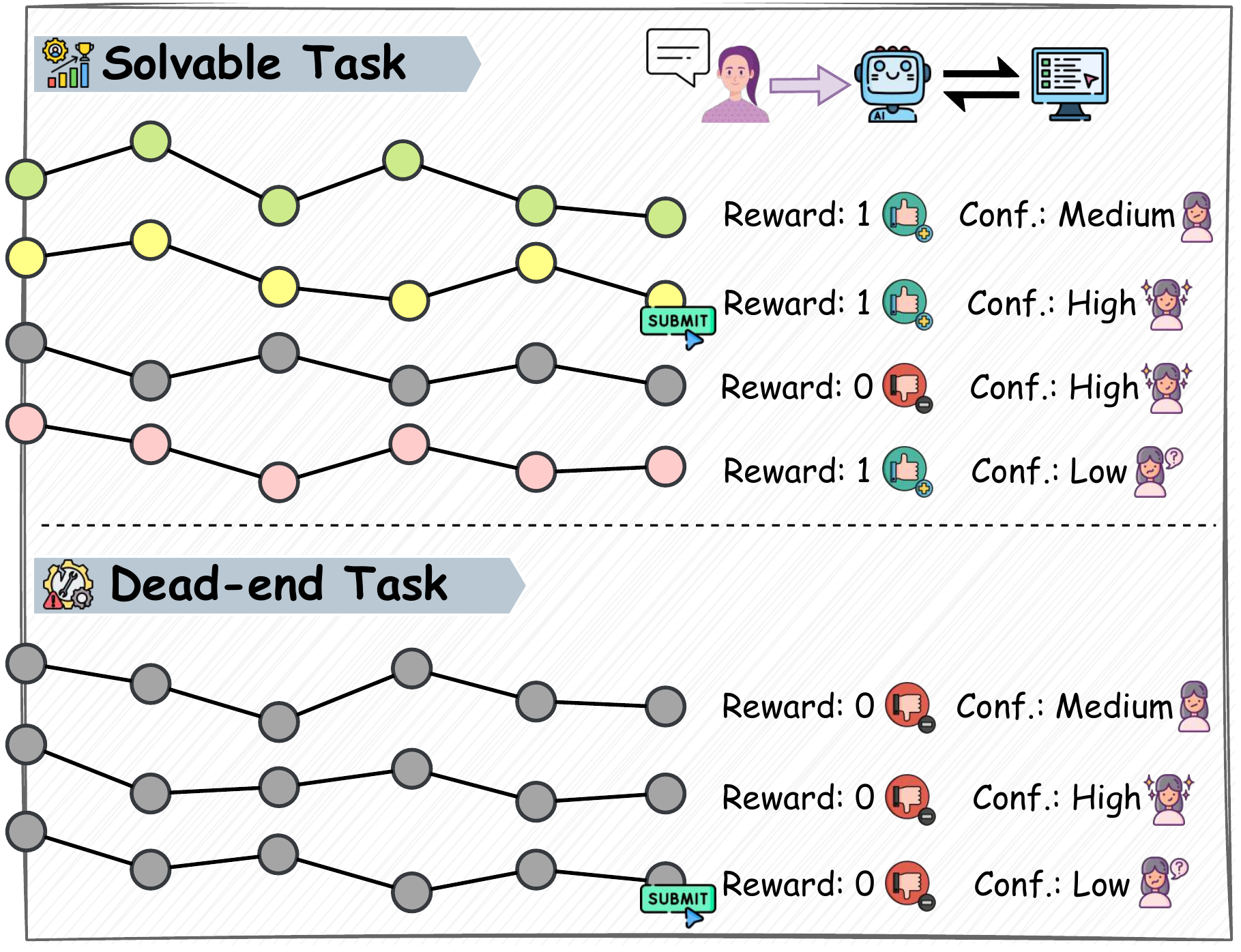}
    \caption{Illustration of the Reward-Guided Scaling strategy.}
    \vspace{-1.6em}
    \label{fig:rewardsup}
\end{wrapfigure} 

Beyond serving as a reward signal for reinforcement learning, VAGEN effectively guides the test-time scaling of the actor agent via a rejection sampling approach (Best-of-$N$) \cite{agents3,prore}, as illustrated in Figure~\ref{fig:rewardsup}. For a given task $q$, we sample $N$ independent trajectories $\{\mathcal{T}_j\}_{j=1}^N$ from the actor agent. The verifier agent $\pi_e$ evaluates each trajectory to obtain a predicted reward and confidence tuple $(\hat{R}_j, \hat{C}_j)$. The final trajectory $\mathcal{T}^*$ is selected based on the verification results: we prioritize trajectories classified as successful ($\hat{R}=1$) and select the one with the highest confidence $\hat{C}$. In the scenario where all trajectories are predicted as failures ($\forall j, \hat{R}_j=0$), we randomly select one candidate for submission. This strategy leverages VAGEN to filter out suboptimal execution paths, thereby enhancing the overall pass rate at inference time. We provide a detailed theoretical analysis of the final SR achievable by the actor agent (without considering confidence) in Appendix~\ref{sec:theorem}. 

\section{Experiments}

This section comprehensively evaluates VAGEN in terms of verification accuracy, cross-environment generalization, online reinforcement learning, test-time scaling, and the effectiveness of its key components across diverse experimental settings.

\subsection{Experimental Settings}

\noindent \textbf{Benchmark.} We conduct our primary evaluation on the OSWorld-Verified benchmark (361 tasks, no Google Drive related tasks), which focuses on desktop operating system environments. To validate the generalization and robustness of VAGEN across platforms, we also conduct experiments on the mobile-based AndroidWorld benchmark. 

\noindent \textbf{Evaluation Protocol.} We employ diverse actor agents to execute benchmark tasks and utilize the benchmark's test scripts and human evaluation to obtain the ground truth regarding task success. Subsequently, a verifier agent provides a predicted judgement. We evaluate binary classification metrics, including precision, recall, F1, and accuracy. We use actor agents with varying SRs under a 50-step budget to verify the performance of VAGEN in both class-balanced and class-imbalanced scenarios. Specifically, these include Claude-Sonnet-4.5 \cite{claude} and UI-TARS-1.5-7B \cite{uitars}, which achieve SRs of 55.9\% and 22.4\%, respectively. See Appendix~\ref{sec:expdetail} for detailed experimental settings. The source code of VAGEN is available at \url{https://github.com/CcQunResearch/VAGEN}.

\noindent \textbf{Baselines.} We compare with the following baselines:
\begin{itemize} 
\item \textbf{DigiRL} \cite{digirl} assesses by feeding the task description and the final screenshot into the VLM.
\item \textbf{DistRL} \cite{distrl} incorporates the last two actions, the task description and final screenshot as input for VLM evaluation.
\item \textbf{WebRL} \cite{webrl} leverages the entire trajectory, providing the VLM with the task description, the final screenshot, and the complete action history.
\item \textbf{AndroidGen} \cite{androidgen} takes the full action history and the final screenshot as input. It decomposes complex task into multiple sub-goals, and the VLM determines success based on whether all these sub-goals have been achieved.
\item \textbf{ZeroGUI} \cite{zerogui} relies exclusively on screenshots from all steps as VLM input. It explicitly excludes textual responses to mitigate potential hallucinations from the actor agent.
\item \textbf{FullTrajEval} evaluates using complete trajectories; specifically, it builds upon ZeroGUI by additionally receiving the full action history.
\item \textbf{ProRe} \cite{prore} adopts a reasoner-actor collaboration architecture to perform active environment probing for verification.
\end{itemize} 

\subsection{Main Results and Discussion}

Table~\ref{tab:osworld} presents the performance of different methods across both actor agents on the OSWorld-Verified benchmark over 50 steps. VAGEN and all baseline methods employ Claude-Sonnet-4.5 as the backbone. Due to context length constraints, ZeroGUI and FullTrajEval utilize screenshots from the last 15 steps. We further observe that the OSWorld-Verified benchmark still suffers from limited ground truth issues caused by false negatives \cite{cuarewardbench,osworld_verified}. This leads to lower precision for the test-script-based ground truth in Table~\ref{tab:osworld}. Therefore, we conduct human evaluation to obtain the actual ground truth.

Table~\ref{tab:osworld} shows that VAGEN comprehensively outperforms other methods in both class-balanced and imbalanced scenarios, achieving an accuracy exceeding 90\% with the human evaluation ground truth. This highlights the clear superiority of the trajectory-grounded interactive verification paradigm over both passive and active verifiers. LLM-as-a-Judge methods exhibit lower F1 scores in class-imbalanced settings, indicating an inadequacy in identifying correct trajectories from weaker actor agents. This deficiency severely restricts the capability to capture sparse high-value signals during RL training. ZeroGUI and FullTrajEval utilize more screenshot information than other baseline methods, resulting in improved performance; this underscores the necessity of deriving evidence from visual signals. Notably, in class-imbalanced scenarios, FullTrajEval's incorporation of textual responses from the actor agent reduces performance compared to ZeroGUI. This suggests that the reasoning content of low-capability actor agents contains significant hallucinations, thereby impeding the judgment of the reward model. ProRe adopts a Reasoner-Actor multi-model collaborative architecture. Compared with VAGEN’s end-to-end verification process, it introduces inter-component communication costs and potential information loss. Moreover, its reasoner model generates probing tasks without direct access to the trajectory or environment, which can lead to hallucinated probing tasks. Additionally, we provide a behavioral analysis and case study of VAGEN in Appendix~\ref{sec:behavioral} and Appendix~\ref{sec:caseatudy}.

\begin{table*}[!h]
\centering
\caption{Evaluation performance on the OSWorld-Verified benchmark across class-balanced and class-imbalanced settings. The 1st and 2nd best results are denoted as \colorbox{mlb}{\textbf{blue}} and \colorbox{mlo}{\textbf{orange}}, respectively.}
\resizebox{\textwidth}{!}{
\begin{tabular}{lcccc|cccc|cccc|cccc}
\Xhline{1.0pt}
 \rowcolor{gray!20}
 ~ & \multicolumn{8}{c}{\textbf{Claude-Sonnet-4.5 (Class-Balanced)}} & \multicolumn{8}{c}{\textbf{UI-TARS-1.5-7B (Class-Imbalanced)}} \\
 \cline{2-17}
 \rowcolor{gray!20}
 ~ & \multicolumn{4}{c}{\textbf{Test Scripts}} & \multicolumn{4}{c}{\textbf{Human Evaluation}} & \multicolumn{4}{c}{\textbf{Test Scripts}} & \multicolumn{4}{c}{\textbf{Human Evaluation}} \\
 \cline{2-17}
 \rowcolor{gray!20}
 \multirow{-3}{*}{\textbf{Method}} & Prec & Rec & F1 & Acc & Prec & Rec & F1 & Acc & Prec & Rec & F1 & Acc & Prec & Rec & F1 & Acc\\
 \hline
 DigiRL & 73.9 & 73.5 & 73.7 & 70.2 & 85.7 & 72.8 & 78.7 & 73.8 & 39.2 & 38.3 & 38.7 & 72.9 & 68.4 & 52.9 & 59.7 & 79.8 \\
 DistRL & 74.9 & 74.1 & 74.5 & 70.9 & 87.7 & 74.2 & 80.4 & 75.6 & 43.4 & 40.7 & 42.0 & 74.8 & 73.7 & 53.8 & 62.2 & 81.2\\
 WebRL & 77.8 & 76.7 & 77.3 & 73.9 & 91.6 & 77.2 & 83.8 & 79.8 & 44.2 & 42.0 & 43.0 & 75.1 & 76.6 & 56.2 & 64.8 & 82.2 \\
 AndroidGen & \colorbox{mlo}{\textbf{78.8}} & 76.6 & 77.7 & 74.4 & \colorbox{mlo}{\textbf{92.6}} & 77.0 & 84.1 & 80.2 & 45.3 & 48.1 & 46.7 & 75.3 & 73.6 & 60.4 & 66.3 & 82.0 \\
 ZeroGUI & 75.1 & 89.2 & 81.5 & 77.2 & 87.1 & 88.2 & 87.7 & 83.7 & 52.3 & 71.6 & 60.4 & 78.9 & 76.6 & 79.4 & 78.0 & 86.7\\
 FullTrajEval & 75.6 & 90.1 & 82.2 & 78.1 & 87.7 & 89.5 & 88.6 & 84.7 & 50.5 & 69.1 & 58.3 & 77.8 & 74.8 & 76.9 & 75.8 & 85.3 \\
 ProRe & 77.4 & \colorbox{mlo}{\textbf{91.4}} & \colorbox{mlo}{\textbf{83.8}} & \colorbox{mlo}{\textbf{80.2}} & 90.0 & \colorbox{mlo}{\textbf{90.8}} & \colorbox{mlo}{\textbf{90.4}} & \colorbox{mlo}{\textbf{87.3}} & \colorbox{mlo}{\textbf{56.0}} & \colorbox{mlo}{\textbf{75.3}} & \colorbox{mlo}{\textbf{64.2}} & \colorbox{mlo}{\textbf{81.2}} & \colorbox{mlo}{\textbf{80.2}} & \colorbox{mlo}{\textbf{81.7}} & \colorbox{mlo}{\textbf{80.9}} & \colorbox{mlo}{\textbf{88.4}}\\
 \hdashline
 VAGEN & \colorbox{mlb}{\textbf{79.2}} & \colorbox{mlb}{\textbf{94.4}} & \colorbox{mlb}{\textbf{86.2}} & \colorbox{mlb}{\textbf{83.1}} &  \colorbox{mlb}{\textbf{94.0}} & \colorbox{mlb}{\textbf{95.2}} & \colorbox{mlb}{\textbf{94.6}} & \colorbox{mlb}{\textbf{92.9}}  & \colorbox{mlb}{\textbf{61.9}} & \colorbox{mlb}{\textbf{86.4}} & \colorbox{mlb}{\textbf{72.2}} & \colorbox{mlb}{\textbf{85.0}} & \colorbox{mlb}{\textbf{88.5}} & \colorbox{mlb}{\textbf{90.1}} & \colorbox{mlb}{\textbf{89.3}} & \colorbox{mlb}{\textbf{93.4}} \\
\Xhline{1.0pt}
\end{tabular}
}
\label{tab:osworld}
\end{table*}

\begin{wraptable}{r}{0.48\textwidth}
\vspace{-2em}
\centering
\caption{Generalization results on the AndroidWorld benchmark.}
\resizebox{0.48\textwidth}{!}{
\begin{tabular}{lcccc|cccc}
\Xhline{1.0pt}
 \rowcolor{gray!20}
 ~ & \multicolumn{4}{c}{\textbf{Claude-Sonnet-4.5}} & \multicolumn{4}{c}{\textbf{UI-TARS-1.5-7B}} \\
 \cline{2-9}
 \rowcolor{gray!20}
 \multirow{-2}{*}{\textbf{Method}} & Prec & Rec & F1 & Acc & Prec & Rec & F1 & Acc \\
 \hline
 DigiRL & 79.1 & 81.5 & 80.3 & 77.6 & 77.4 & 68.6 & 72.7 & 84.5 \\
 DistRL & 82.9 & 89.2 & 85.9 & 83.6 & 76.5 & 74.3 & 75.4 & 85.3 \\
 WebRL & 83.8 & 87.7 & 85.7 & 83.6 & 75.7 & 80.0 & 77.8 & 86.2 \\
 AndroidGen & 85.3 & 89.2 & 87.2 & 85.3 & 76.9 & \colorbox{mlo}{\textbf{85.7}} & 81.1 & 87.9 \\
 ZeroGUI & 86.8 & 90.8 & 88.7 & 87.1 & 80.6 & 82.9 & 81.7 & 88.8 \\
 FullTrajEval & 88.1 & 90.8 & 89.4 & 87.9 & 76.3 & 82.9 & 79.5 & 87.1 \\
 ProRe & \colorbox{mlo}{\textbf{91.0}} & \colorbox{mlo}{\textbf{93.8}} & \colorbox{mlo}{\textbf{92.4}} & \colorbox{mlo}{\textbf{91.4}} & \colorbox{mlo}{\textbf{85.3}} & 82.9 & \colorbox{mlo}{\textbf{84.1}} & \colorbox{mlo}{\textbf{90.5}}\\
 \hdashline
 VAGEN & \colorbox{mlb}{\textbf{92.5}} & \colorbox{mlb}{\textbf{95.4}} & \colorbox{mlb}{\textbf{93.9}} & \colorbox{mlb}{\textbf{93.1}} & \colorbox{mlb}{\textbf{86.1}} & \colorbox{mlb}{\textbf{88.6}} & \colorbox{mlb}{\textbf{87.3}} & \colorbox{mlb}{\textbf{92.2}} \\
\Xhline{1.0pt}
\end{tabular}
}
\label{tab:androidworld}
\end{wraptable}

We further verify the effectiveness of VAGEN on the AndroidWorld benchmark in Table~\ref{tab:androidworld}, demonstrating the scalability and robustness of our method across different environments. The results show that VAGEN achieves strong performance. Since the AndroidWorld mobile environment is relatively simpler, ProRe is less affected by reasoner hallucinations and thus also achieves good performance on this benchmark.

\subsection{Online Reinforcement Learning with Different Reward Models}

\begin{wrapfigure}{r}{0.49\textwidth}
    \vspace{-1.3em}
    \includegraphics[width=0.49\textwidth]{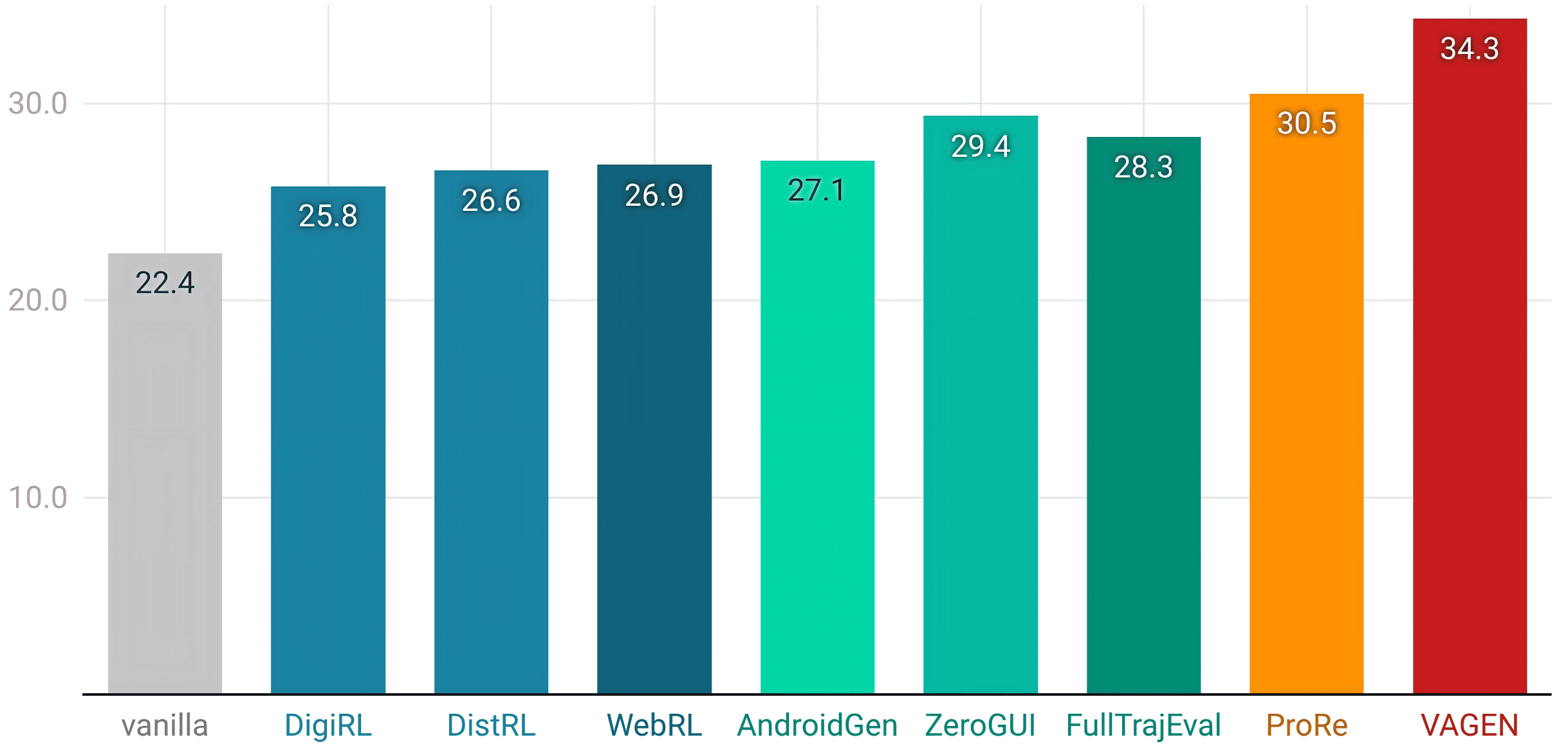}
    \caption{OSWorld-Verified success rates after GRPO training with different reward models.}
    \vspace{-0.5em}
    \label{fig:rl}
\end{wrapfigure} 

Using all 6K AgentSynth \cite{agentsynth} tasks for online RL, we compare different reward models under UI-TARS-1.5-7B backbone and GRPO \cite{grpo} training setup, where each task is sampled with 16 rollouts and the final policy is evaluated on OSWorld-Verified. See Appendix~\ref{sec:expdetailmain} for detailed settings. As shown in Figure~\ref{fig:rl}, all reward models improve upon the vanilla baseline of 22.4\%, confirming that synthetic OSWorld tasks can provide useful online RL experience. Stronger trajectory-aware rewards, such as ZeroGUI and FullTrajEval, outperform DigiRL, DistRL, and WebRL, indicating the importance of richer verification evidence. VAGEN achieves 34.3\% of SR, surpassing the strongest baseline ProRe by 3.8 points. These results show that VAGEN's proactive verification produces more accurate and actionable reward signals for policy optimization.

\subsection{Test-Time Scaling}

\subsubsection{Scaling of Verifier Agent}

\begin{wrapfigure}{r}{0.49\textwidth}
    \vspace{-0.2em}
    \includegraphics[width=0.51\textwidth]{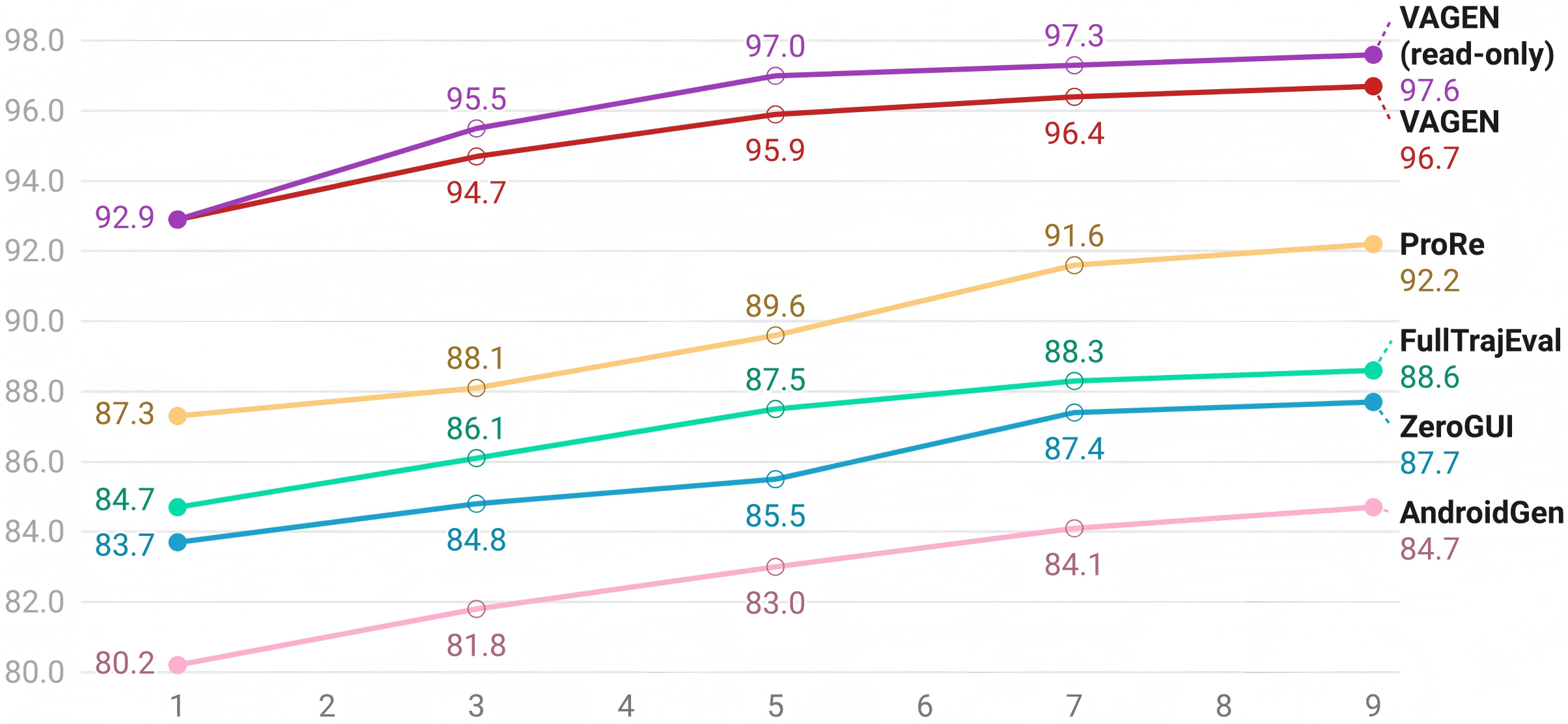}
    \caption{Read-only scaling for verifier agent (Claude-Sonnet-4.5).}
    \vspace{-1.2em}
    \label{fig:readonly}
\end{wrapfigure} 

Figure~\ref{fig:readonly} shows the read-only scaling experiments (accuracy reported) for the verifier agent introduced in Section~\ref{sec:readonly} (using human evaluation ground truth). We scale ProRe by replaying the trajectory. We perform an odd number of verifications and apply a majority vote. The verifier agent is restricted from performing operations that alter key verification evidence (e.g., deleting or writing files, changing settings). Detailed settings are shown in Appendix~\ref{sec:readonlysetting}. The results indicate that VAGEN achieves the most significant performance gain from scaling under read-only constraints. This is likely because VAGEN relies on multi-step verification via multiple tools, offering a broader range of verification paths compared to other methods, while interaction with the environment ensures access to complete verification evidence. Even without the read-only restriction, VAGEN achieves obvious performance improvements, suggesting that the verification task itself does not require extensive write operations that would alter key evidence. Note that due to the prohibitive cost of human evaluation, we rely on script-based ground truth for the subsequent experiments.

\subsubsection{Scaling of Actor Agent}

\begin{figure*}[!h]
  \centering
  \subfigure[Claude-Sonnet-4.5\label{fig:s1}]{\includegraphics[width=0.48\textwidth]{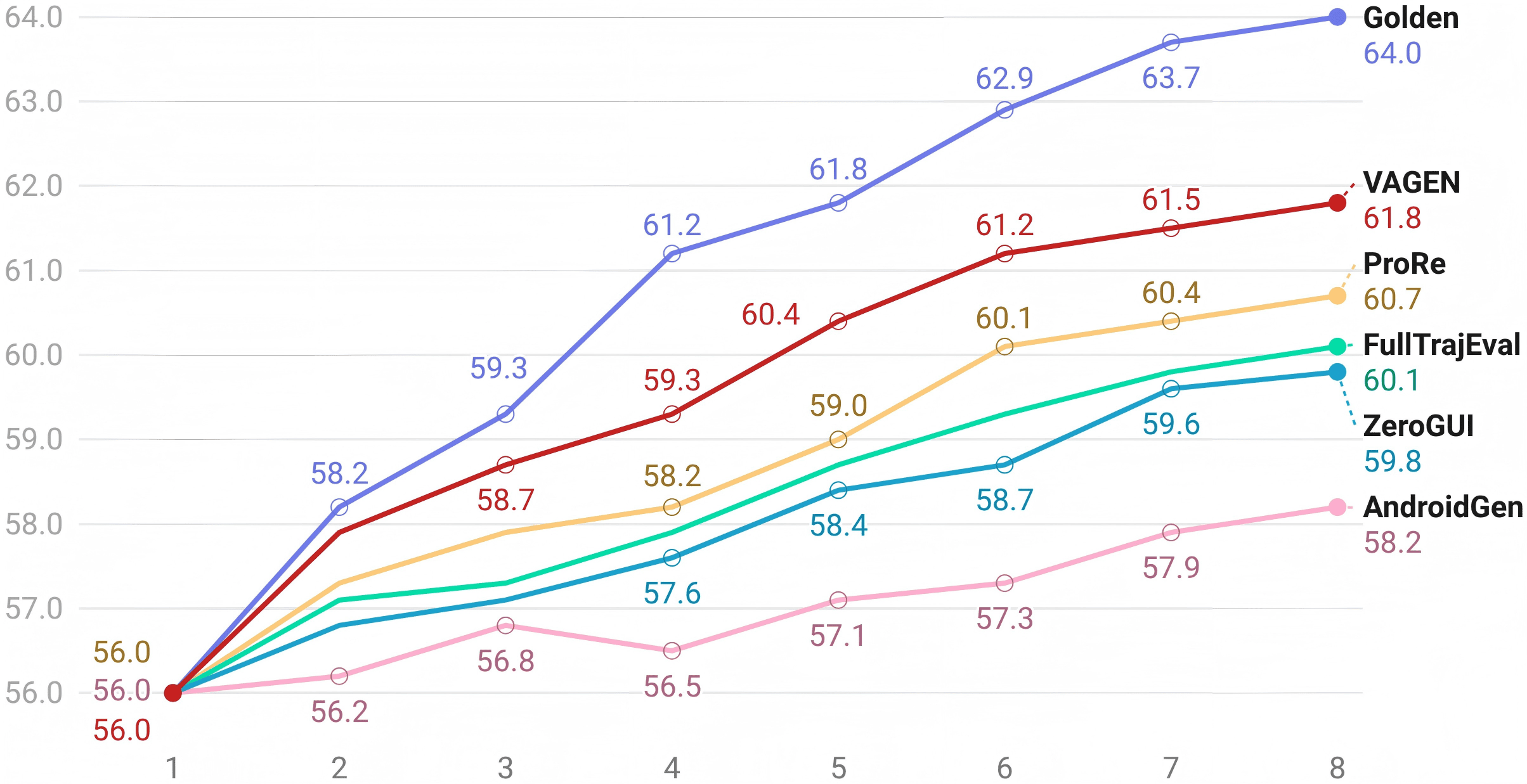}}
  \subfigure[UI-TARS-1.5-7B\label{fig:s2}]{\includegraphics[width=0.48\textwidth]{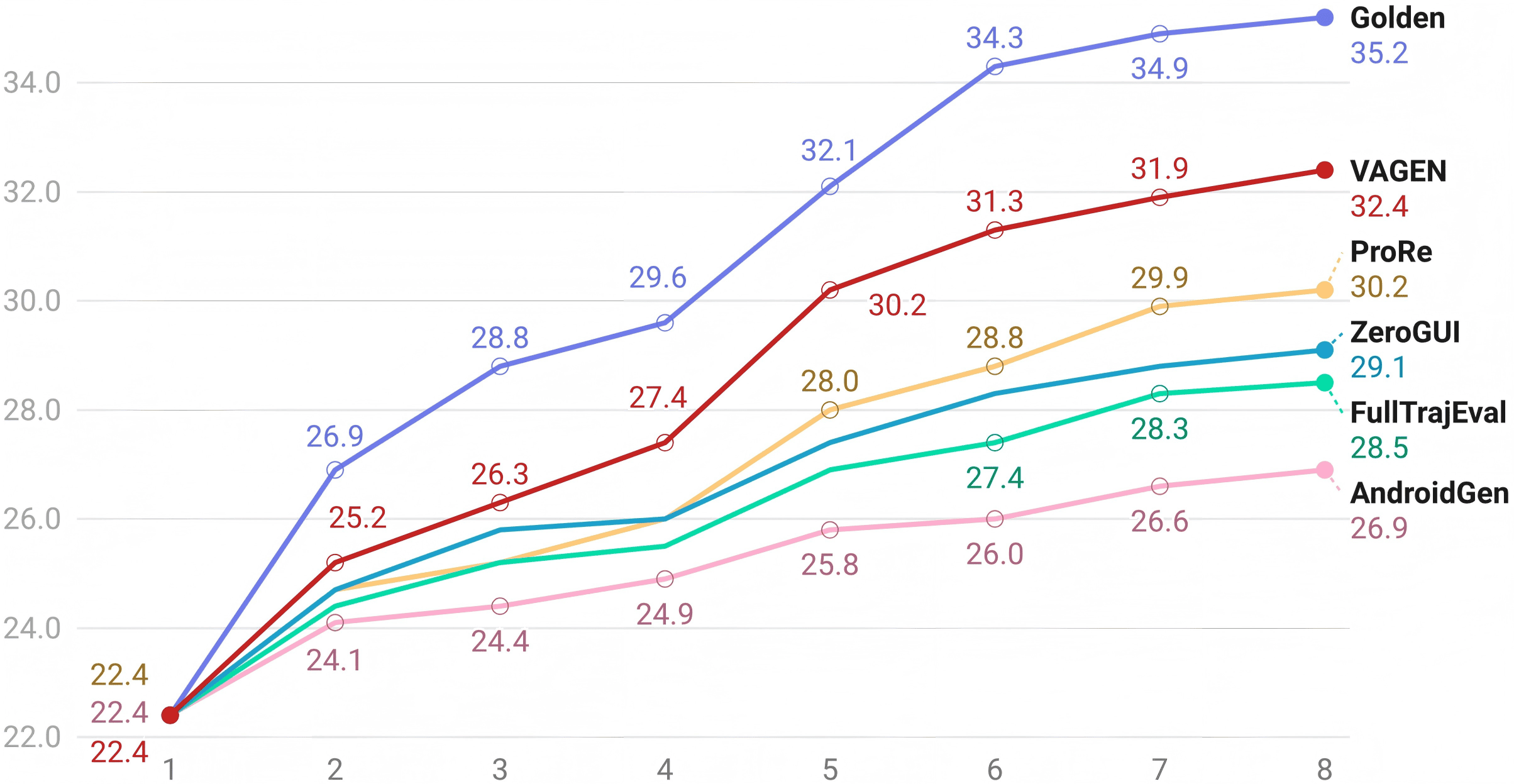}}
  \caption{Results of reward-guided scaling for the actor agent using Best-of-N rejection sampling.}
  \label{fig:rewardguided}
\end{figure*}

The most crucial role of a good reward model is to guide the actor agent in performing more efficient rollouts, thereby achieving superior performance during RL training or inference time. Figure~\ref{fig:rewardguided} presents the reward-guided scaling experiments for the actor agent introduced in Section~\ref{sec:rewardguided} (reporting SR). The results show that VAGEN achieves the most significant improvement compared to other baseline methods, indicating its ability to provide more reliable rewards for actor agent rollouts. VAGEN matches the performance of baseline methods at $N=8$ with just $N=5$, demonstrating the superior sample efficiency. Furthermore, VAGEN predicts successful rollout trajectories with higher precision, which contributes to resilience to false positives during the RL training process.

\subsection{Ablation Study}

We ablate VAGEN's key components in Table~\ref{tab:ab}.
VAGEN outperforms ZeroGUI and ProRe while requiring fewer screenshots and avoiding excessive output token consumption, demonstrating the rationality of VAGEN's overall design.
Memory Consolidation positively impacts performance for UI-TARS-1.5-7B because the reasoning of low-capability actor agents contains numerous hallucinations and errors, whereas Memory Consolidation retains only factual agent operations. 
Removing the Progressive Verification Mechanism forces the agent to plan verification paths completely autonomously, leading to a significant performance decline. This suggests the mechanism provides prior guidance for the effective utilization of the four tools. 
Without \textit{check screenshot}, VAGEN’s ability to obtain direct evidence from the trajectory is weakened, leading to performance degradation. This highlights the importance of cues in the trajectory. In ProRe, however, the trajectory is reduced to a reference for consistency checking against probing results, rather than serving as the primary basis for judgment. Moreover, ProRe follows a relatively fixed verification pipeline, which overlooks the heterogeneous difficulty distribution of tasks.
We also ablate the three tools used for environment interaction, restricting the verifier to rely solely on trajectories. This achieves performance comparable to the LLM-as-a-Judge method but with significantly lower resource consumption, as the agent can autonomously select specific screenshots for inspection based on operations $\mathcal{H}$. 
Further ablation of specific tools reveals that the direct interaction capability afforded by the \textit{computer use} tool is key to VAGEN's superior performance, while the \textit{execute shell} and \textit{execute python} tools serve as auxiliary means for indirect interaction to reduce resource consumption.
We also investigate the effects of different backbone models and inference budgets in Appendix~\ref{sec:extexp}.

\begin{table*}[!h]
\centering
\caption{Ablation study of VAGEN components. \colorbox{mlb}{\textbf{Blue}} and \colorbox{mlo}{\textbf{orange}} mark significant gains and drops relative to VAGEN. `\# img/token' is the counts of input images (screenshots) and output tokens.}
\resizebox{\textwidth}{!}{
\begin{tabular}{lcccc|cccc}
\Xhline{1.0pt}
 \rowcolor{gray!20}
 ~ & \multicolumn{4}{c}{\textbf{Claude-Sonnet-4.5}} & \multicolumn{4}{c}{\textbf{UI-TARS-1.5-7B}} \\
 \cline{2-9}
 \rowcolor{gray!20}
 ~ & F1 (↑) & Acc (↑) & Avg Steps (↓) & \# img/token (↓) & F1 (↑) & Acc (↑) & Avg Steps (↓) & \# img/token (↓) \\
 \hline
 ZeroGUI & 81.5 & 77.2 & - & 14.1/832 & 60.4 & 78.9 & - & 14.6/884\\
 ProRe & 83.8 & 80.2 & 25.2 & 24.2/1923 & 64.2 & 81.2 & 24.6 & 23.6/1884 \\
 VAGEN & 86.2 & 83.1 & 17.4 & 11.9/1218 & 72.2 & 85.0 & 15.3 & 10.9/1104 \\
 \hdashline
 w/o Memory Consolidation & 85.9 & 82.8 & 17.7 & 12.2/1314 & \colorbox{mlo}{\textbf{69.9}} & \colorbox{mlo}{\textbf{82.8}} & 16.2 & 11.5/1193 \\
 w/o Progressive Verification Mechanism & \colorbox{mlo}{\textbf{84.5}} & \colorbox{mlo}{\textbf{80.8}} & 16.7 & 13.4/1137 & \colorbox{mlo}{\textbf{68.0}} & \colorbox{mlo}{\textbf{82.5}} & \colorbox{mlo}{\textbf{17.3}} & \colorbox{mlo}{\textbf{12.7}}/\colorbox{mlo}{\textbf{1256}} \\
 w/o check screenshot tool & \colorbox{mlo}{\textbf{84.3}} & \colorbox{mlo}{\textbf{80.5}} & \colorbox{mlo}{\textbf{21.3}} & \colorbox{mlo}{\textbf{18.4}}/\colorbox{mlo}{\textbf{1603}} & \colorbox{mlo}{\textbf{67.5}} & \colorbox{mlo}{\textbf{82.2}} & \colorbox{mlo}{\textbf{20.7}} & \colorbox{mlo}{\textbf{17.4}}/\colorbox{mlo}{\textbf{1647}}\\
 w/o environment interaction tools & \colorbox{mlo}{\textbf{82.8}} & \colorbox{mlo}{\textbf{78.5}} & \colorbox{mlb}{\textbf{6.4}} & \colorbox{mlb}{\textbf{5.4}}/\colorbox{mlb}{\textbf{532}} & \colorbox{mlo}{\textbf{63.4}} & \colorbox{mlo}{\textbf{79.2}} & \colorbox{mlb}{\textbf{6.9}} & \colorbox{mlb}{\textbf{5.9}}/\colorbox{mlb}{\textbf{582}} \\
 \quad w/o execute shell \& execute python & 85.3 & 82.0 & \colorbox{mlo}{\textbf{19.7}} & \colorbox{mlo}{\textbf{18.7}}/\colorbox{mlo}{\textbf{1477}} & 71.1 & 84.5 & \colorbox{mlo}{\textbf{18.6}} & \colorbox{mlo}{\textbf{17.6}}/\colorbox{mlo}{\textbf{1507}}\\
 \quad w/o computer use & \colorbox{mlo}{\textbf{83.4}} & \colorbox{mlo}{\textbf{79.4}} & \colorbox{mlb}{\textbf{7.2}} & \colorbox{mlb}{\textbf{4.9}}/\colorbox{mlb}{\textbf{602}} & \colorbox{mlo}{\textbf{66.0}} & \colorbox{mlo}{\textbf{80.9}} & \colorbox{mlb}{\textbf{8.6}} & \colorbox{mlb}{\textbf{6.9}}/\colorbox{mlb}{\textbf{693}}\\
 \Xhline{1.0pt}
\end{tabular}
}
\label{tab:ab}
\end{table*}

\section{Conclusion}

We revisit reward modeling for GUI agents and identify the complementary strengths of trajectory verification and environment interaction. To unify both, we propose VAGEN, a trajectory-grounded interactive verification framework instantiated via a tool-augmented verifier agent governed by a Progressive Verification Mechanism. Extensive experiments demonstrate that VAGEN significantly improves evaluation accuracy with a favorable performance-efficiency trade-off.

\bibliographystyle{plainnat}
\bibliography{reference}

\clearpage
\appendix

\section*{Appendix Contents}

\begin{table}[!h]
    \centering
    \footnotesize
    \begin{tabular}{cl}
    \textbf{Appendix Sections} & \textbf{Contents} \\ \toprule
    \autoref{sec:expdetail} & \begin{tabular}[c]{@{}l@{}} More Details on Experimental Settings \end{tabular} \\ \midrule
    \autoref{sec:extexp} & \begin{tabular}[c]{@{}l@{}} Extended Evaluation Experiments \end{tabular} \\ \midrule
    \autoref{sec:behavioral} & \begin{tabular}[c]{@{}l@{}} Behavioral Analysis of Verifier Agent \end{tabular} \\ \midrule
    \autoref{sec:theorem} & \begin{tabular}[c]{@{}l@{}} Theory: Reward-Guided Scaling of Actor Agent \end{tabular} \\ \midrule
    \autoref{sec:discussion} & \begin{tabular}[c]{@{}l@{}} Further Discussion on VAGEN \end{tabular} \\ \midrule
    \autoref{sec:pi} & \begin{tabular}[c]{@{}l@{}} Prompts and Case Study of VAGEN \end{tabular} \\
    \bottomrule
    \end{tabular}    
\end{table}

\section{More Details on Experimental Settings}
\label{sec:expdetail}

In this section, we introduce the main settings used in the experiments of this paper. Except for certain experiments with special hyperparameters, the same hyperparameters are adopted to ensure consistency and fairness in experimental comparisons.

\subsection{Main Setting}
\label{sec:expdetailmain}

We implement our method and baselines primarily using the PyTorch\footnote{\url{https://github.com/pytorch/pytorch}}, openai\footnote{\url{https://github.com/openai/openai-python}}, and vllm\footnote{\url{https://github.com/vllm-project/vllm}} libraries. We conduct our experiments mainly on a single node equipped with eight NVIDIA H100 80GB GPUs. To ensure fairness and consistency in comparisons, we maintain uniform parameter settings across the main experiments and ablation studies. All key hyperparameter configurations are presented in Table~\ref{tab:hp}.

\begin{table}[!h]
\centering
\caption{Hyperparameter configuration in experiments.}
\begin{tabular}{cccl}
 \Xhline{1.0pt}
 \rowcolor{gray!20}
 \textbf{Model} & \textbf{Hyperparameter} & \textbf{Value} & \textbf{Remark} \\
 \hline
 \multirow{5}{*}{\textbf{Actor}} & Temperature & 0.8 & - \\
 ~ & Top-$k$ & 40 & - \\
 ~ & Top-$p$ & 0.9 & - \\
 ~ & Last n screenshots & 10 & In order to control the context \\
 ~ & Screen size & (1280, 720) & Resize from (1920, 1080) to (1280, 720) \\
 \hline
 \multirow{6}{*}{\textbf{Verifier}} & Temperature & 0.8 & - \\
 ~ & Top-$k$ & 40 & - \\
 ~ & Top-$p$ & 0.9 & - \\
 ~ & Last n screenshots & 10 & Prioritize retaining the screenshots in the trajectory\\
 ~ & Max steps & 30 & - \\
 ~ & Screen size & (1280, 720) & Resize from (1920, 1080) to (1280, 720) \\
 \Xhline{1.0pt}
\end{tabular}
\label{tab:hp}
\end{table}

We strictly reimplement the baselines following their original prompts and pipelines. We utilize paid APIs provided by Anthropic, Google, and ByteDance to access models such as Claude-Sonnet-4.5, Gemini-3-Flash, and Doubao-Seed-1.8.

For the online reinforcement learning experiments, we use UI-TARS-1.5-7B as the policy backbone and train it on the 6K AgentSynth tasks with GRPO. For each task, the actor agent samples 16 independent rollouts under a maximum interaction budget of 50 steps. Each compared reward model provides a terminal binary reward for the completed trajectory. We optimize the policy with AdamW using a learning rate of $1\times10^{-6}$, cosine decay with 3\% warmup, gradient clipping at 1.0, a GRPO clipping range of 0.2, and a KL coefficient of 0.01. The main hyperparameters are summarized in Table~\ref{tab:main_hyperparams}. VAGEN confidence is not used during policy optimization and is only used for reward-guided test-time scaling.

\begin{table}[t]
\centering
\caption{Main RL hyperparameter configurations.}
\label{tab:main_hyperparams}
\begin{tabular}{lll}
\Xhline{1.0pt}
\rowcolor{gray!20}
\textbf{Component} & \textbf{Hyperparameter} & \textbf{Value} \\
\hline
\multirow{5}{*}{\textbf{Actor rollout}} & Temperature & 0.8 \\
~ & Top-$k$ / Top-$p$ & 40 / 0.9 \\
~ & Maximum interaction steps & 50 \\
~ & Retained screenshots & 10 \\
~ & Screen resolution & $1280 \times 720$ \\
\hline
\multirow{11}{*}{\textbf{Online RL}} & Rollouts per task & 16 \\
~ & Optimizer & AdamW \\
~ & Learning rate & $1 \times 10^{-6}$ \\
~ & Learning-rate schedule & Cosine decay with 3\% warmup \\
~ & Adam $\beta_1, \beta_2$ & 0.9, 0.95 \\
~ & Weight decay & 0.0 \\
~ & Gradient clipping & 1.0 \\
~ & GRPO clip range & 0.2 \\
~ & KL coefficient & 0.01 \\
~ & Training epoch & 1 \\
~ & Prompt batch size & 8 \\
\Xhline{1.0pt}
\end{tabular}
\end{table}

\subsection{Human Evaluation}

\begin{wrapfigure}{r}{0.45\textwidth}
    \vspace{-1.3em}
    \includegraphics[width=0.45\textwidth]{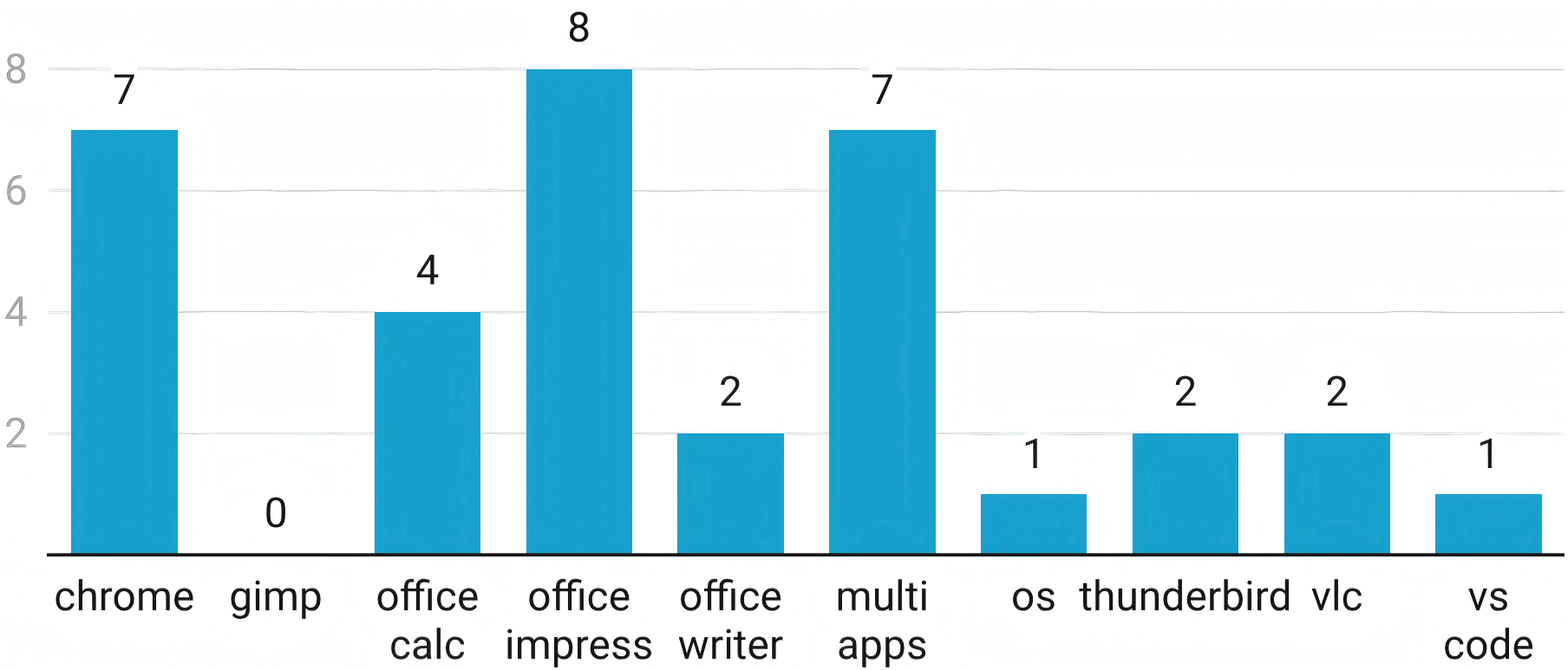}
    \caption{False negative domain distribution.}
    \vspace{-1.8em}
    \label{fig:error}
\end{wrapfigure} 

We acquire human evaluation ground truth through a dual-annotator consistency assessment. We first perform a complete manual audit of all trajectories judged as successful by the test scripts, and find no false positives. This suggests that the positive labels produced by the benchmark test scripts are reliable in our setting. In contrast, the "false negatives" issue in the OSWorld-Verified benchmark primarily stems from multiple task completion paths and ambiguous task descriptions, causing some successful trajectories to be incorrectly rejected by the scripts \cite{osworld_verified}. Consequently, we conduct subsequent manual evaluation exclusively on trajectories predicted as failures by the test scripts. Specifically, if two annotators provide consistent assessments for a task, we accept that result. If their assessments differ, we default to the benchmark’s test script result. Both evaluators are researchers in the field of GUI agents, possessing the requisite professional expertise. Figure~\ref{fig:error} presents the false negative domain distribution with Claude-Sonnet-4.5 as the actor agent. Two annotators identify 37 and 35 false negative samples, respectively, with 34 consistent annotations, resulting in a Jaccard coefficient (inter-annotator agreement rate) of 89.5\%. Human evaluation addresses concerns regarding the low precision in Table~\ref{tab:osworld}, demonstrating that this stems from the unreliability of ground truths in the OSWorld-Verified benchmark rather than verifier bias.

The "false negatives" issue in the OSWorld-Verified test script arises from multiple causes. For instance, it may involve the following types of situations:
\begin{itemize} 
\item The test script presupposes only a single accepted solution, whereas tasks within the complex Ubuntu desktop environment often admit multiple solutions. For example, task@3720f614-37fd-4d04-8a6b-76f54f8c222d requires changing the Chrome browser language to Korean. The agent completes this via shell commands instead of GUI operations, resulting in a failure judgment. 
\item The test script defines incomplete valid output formats for open-ended tasks. For example, task@ce2b64a2-ddc1-4f91-8c7d-a88be7121aac requires the agent to identify a photo of Mount Hua and rename the file using the mountain's name. The agent saves it as "Mount Huashan.jpg", which falls outside the preset allowed names (such as "Huashan Mountain.jpg", "Mount Hua.jpg", and "Hua.jpg"). 
\item Ambiguous task descriptions. For example, task@21ab7b40-77c2-4ae6-8321-e00d3a086c73 requests adding a column to a table containing columns A and B. The agent adds the column at position D, but the test script exclusively verifies column C. 
\end{itemize}

\subsection{Read-Only Restriction}
\label{sec:readonlysetting}

We primarily restrict the verifier agent from performing write operations that alter key verification evidence during read-only scaling through prompt guidance and secondary LLM determination. The specific prompt is detailed in Appendix~\ref{sec:vap}. Figure~\ref{fig:readonly} demonstrates that VAGEN achieves evaluation performance only slightly inferior to read-only scaling even without restrictions. This indicates that VAGEN's verification process predominantly consists of read operations and rarely alters key evidence, even when unrestricted. Nevertheless, we recommend deploying such verifier agents in sandboxed environments with limited permissions. In real-world settings, "read-only" protocols should be implemented to prevent irreversible changes to the system state.

\section{Extended Experiments}
\label{sec:extexp}

In this section, we conduct additional extended experiments to examine the effects of several factors on VAGEN's performance.

\subsection{Backbone Model}

\begin{wraptable}{r}{0.48\textwidth}
\vspace{-1.8em}
\centering
\caption{Performance comparison across different backbones.}
\resizebox{0.48\textwidth}{!}{
\begin{tabular}{lcc|cc}
\Xhline{1.0pt}
 \rowcolor{gray!20}
 ~ & \multicolumn{2}{c}{\textbf{Claude-Sonnet-4.5}} & \multicolumn{2}{c}{\textbf{UI-TARS-1.5-7B}} \\
 \cline{2-5}
 \rowcolor{gray!20}
 \multirow{-2}{*}{\textbf{Model}} & F1 & Acc & F1 & Acc \\
 \hline
 Claude-Sonnet-4.5 & 86.2 & 83.1 & 72.2 & 85.0\\
 Gemini-3-Flash & 85.3 & 82.0 & 72.9 & 84.4\\
 Doubao-Seed-1.8 & 82.1 & 80.9 & 72.3 & 83.9\\
 \Xhline{1.0pt}
\end{tabular}
}
\vspace{-0.5em}
\label{tab:backbone}
\end{wraptable}

To evaluate the generalization and robustness of VAGEN across different backbones, we conduct ablation studies by replacing the base model of verifier agent. We compare the default Claude-Sonnet-4.5 against Gemini-3-Flash\cite{gemini3flash} and Doubao-Seed-1.8\cite{seed18}, with detailed results presented in Table~\ref{tab:backbone}. The results indicate that different backbones achieve substantial evaluation performance, suggesting that VAGEN's critical designs are robust to backbone variations. This demonstrates that its effectiveness does not rely solely on a specific proprietary model but represents a general paradigm transferable to various high-performance LLMs. This also implies that as the capabilities of reasoning models continue to improve, the performance of trajectory-grounded interactive verification is poised for further enhancement.

\subsection{Inference Budget}

\begin{wrapfigure}{r}{0.4\textwidth}
    \vspace{-1.3em}
    \includegraphics[width=0.4\textwidth]{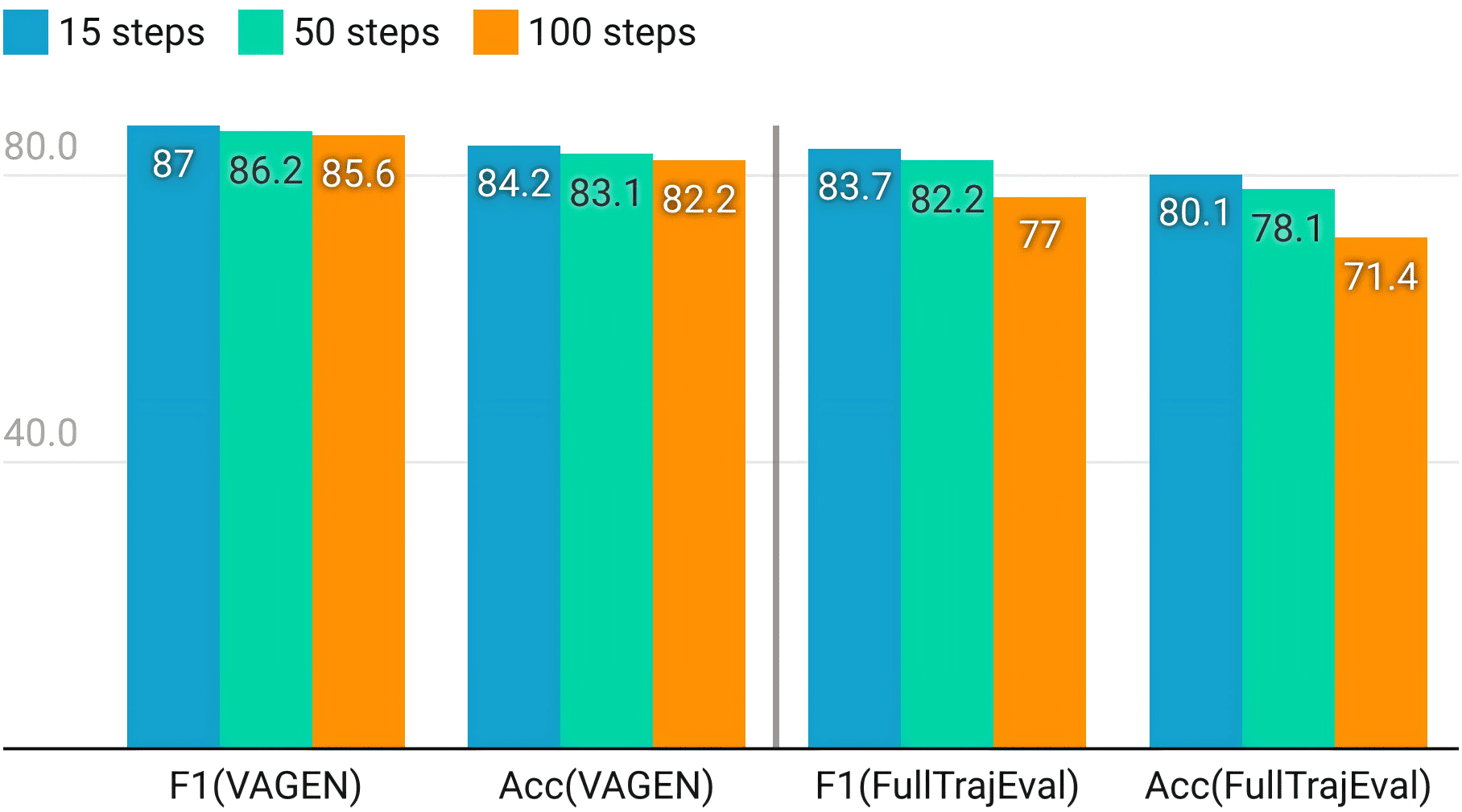}
    \caption{Evaluation performance under varying inference step budgets (Claude-Sonnet-4.5).}
    \vspace{-0.5em}
    \label{fig:steps}
\end{wrapfigure} 

We compare VAGEN and FullTrajEval across different reasoning step budgets of actor agent in Figure~\ref{fig:steps}. The results indicate that FullTrajEval experiences a significant performance decline as trajectory complexity increases, primarily because the LLM-as-a-Judge approach relies excessively on the trajectory itself. In contrast, VAGEN maintains stable performance across varying step budgets. This is mainly because VAGEN, while considering reasoning traces and visual observations within the trajectory, can also probe the final state of the environment for judgment, thereby mitigating the impact of increasing trajectory complexity.

\section{Verifier Behavioral Analysis}
\label{sec:behavioral}

In this section, we statistically analyze the verifier agent's behaviors to intuitively demonstrate how VAGEN acquires critical evidence and completes verification via progressive trajectory-grounded verification with selective environment interaction for more reliable reward modeling.

\subsection{Domain Performance}

Figure~\ref{fig:b1} shows the performance across each domain of the OSWorld-Verified benchmark using human evaluation ground truth. The results demonstrate that VAGEN achieves consistent performance across various software environments. This consistency proves that the proactive verification paradigm possesses robust generalization capabilities and is not limited to specific types of GUI tasks.

\subsection{Tool Invocation Statistics}

\begin{figure}[!h]
  \centering
  \subfigure[Performance across different domains.\label{fig:b1}]{\includegraphics[width=0.398\textwidth]{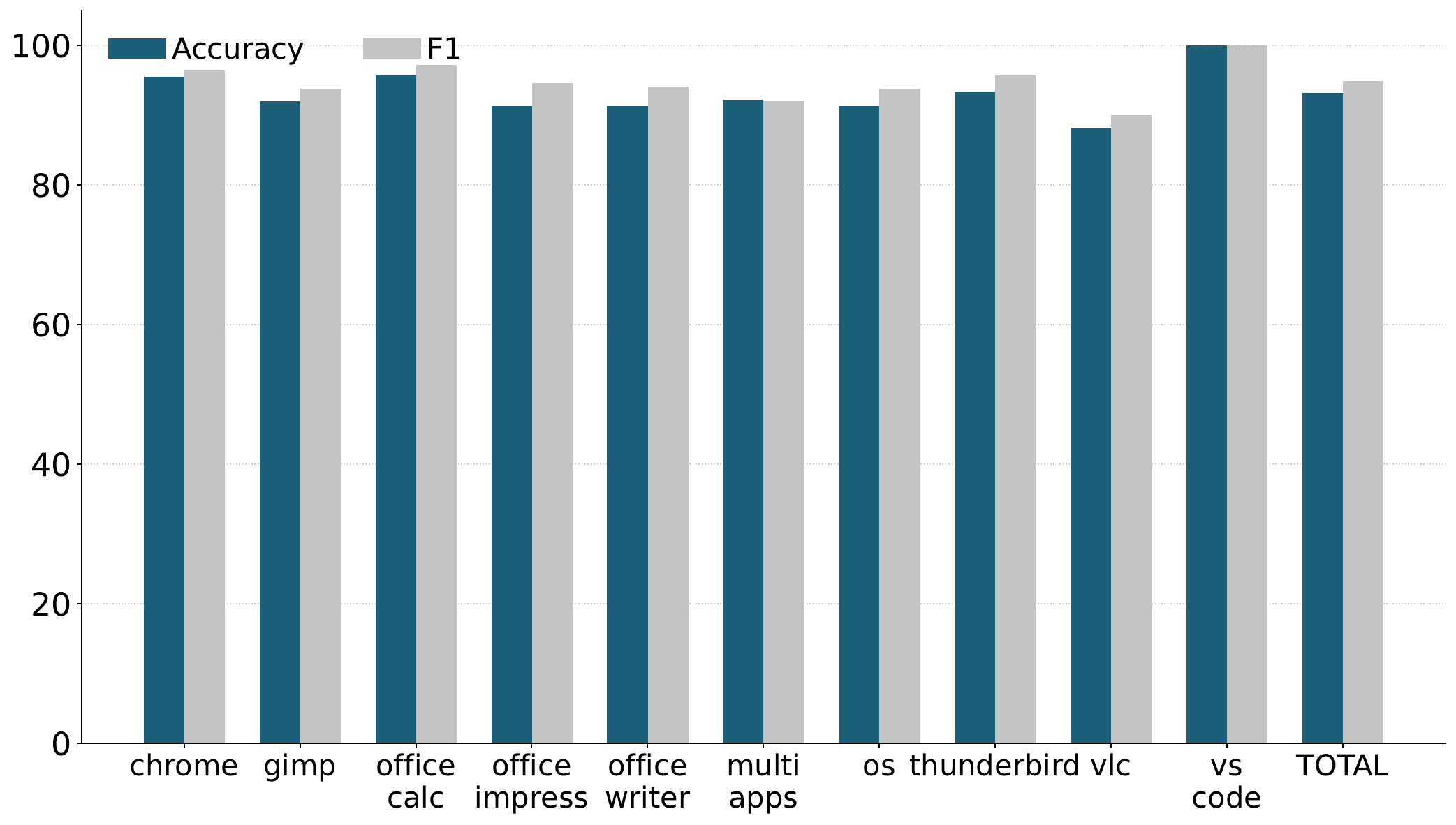}}
  \subfigure[Distribution of tasks requiring specific verification tools.\label{fig:b2}]{\includegraphics[width=0.258\textwidth]{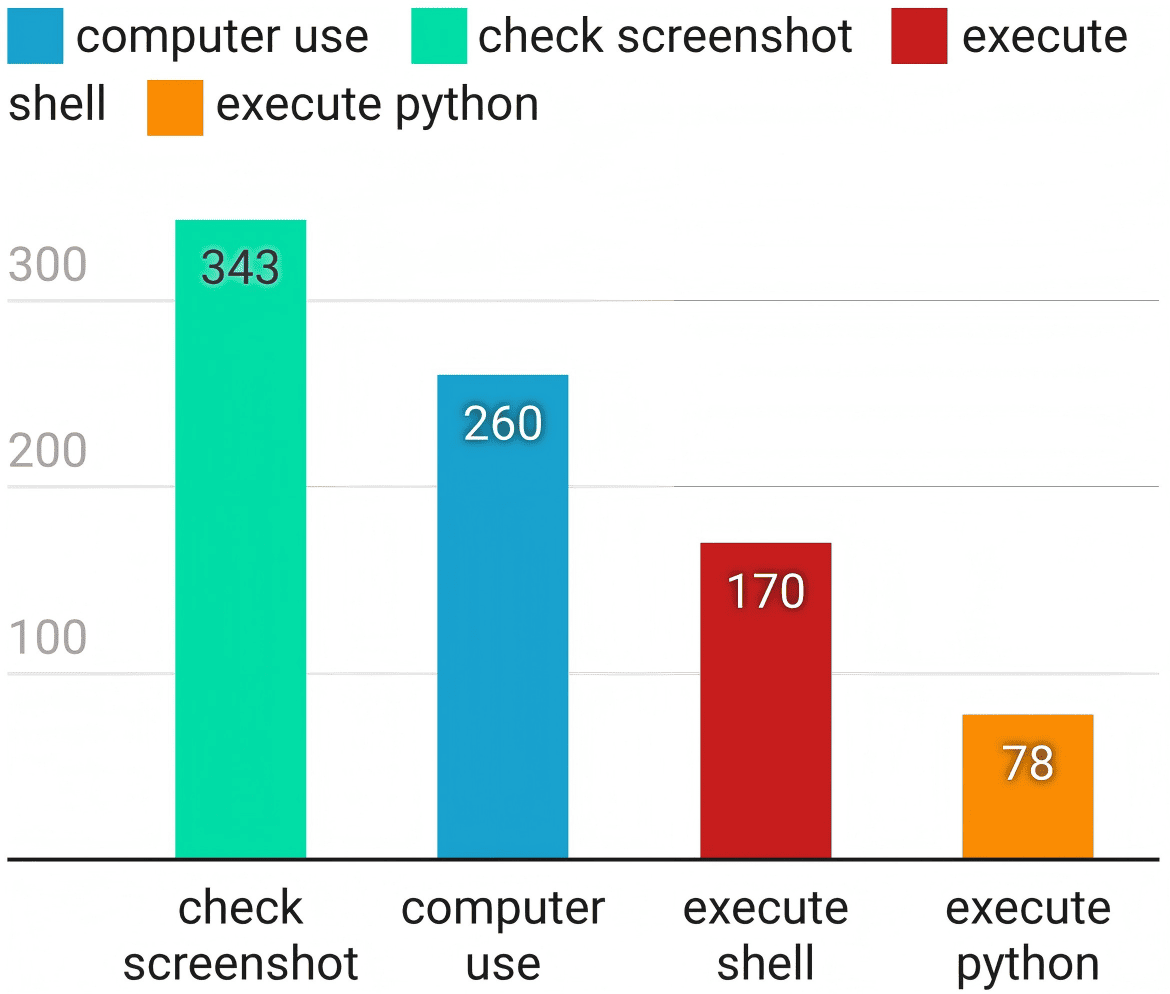}}
  \subfigure[Statistics of total tool invocation frequency.\label{fig:b3}]{\includegraphics[width=0.327\textwidth]{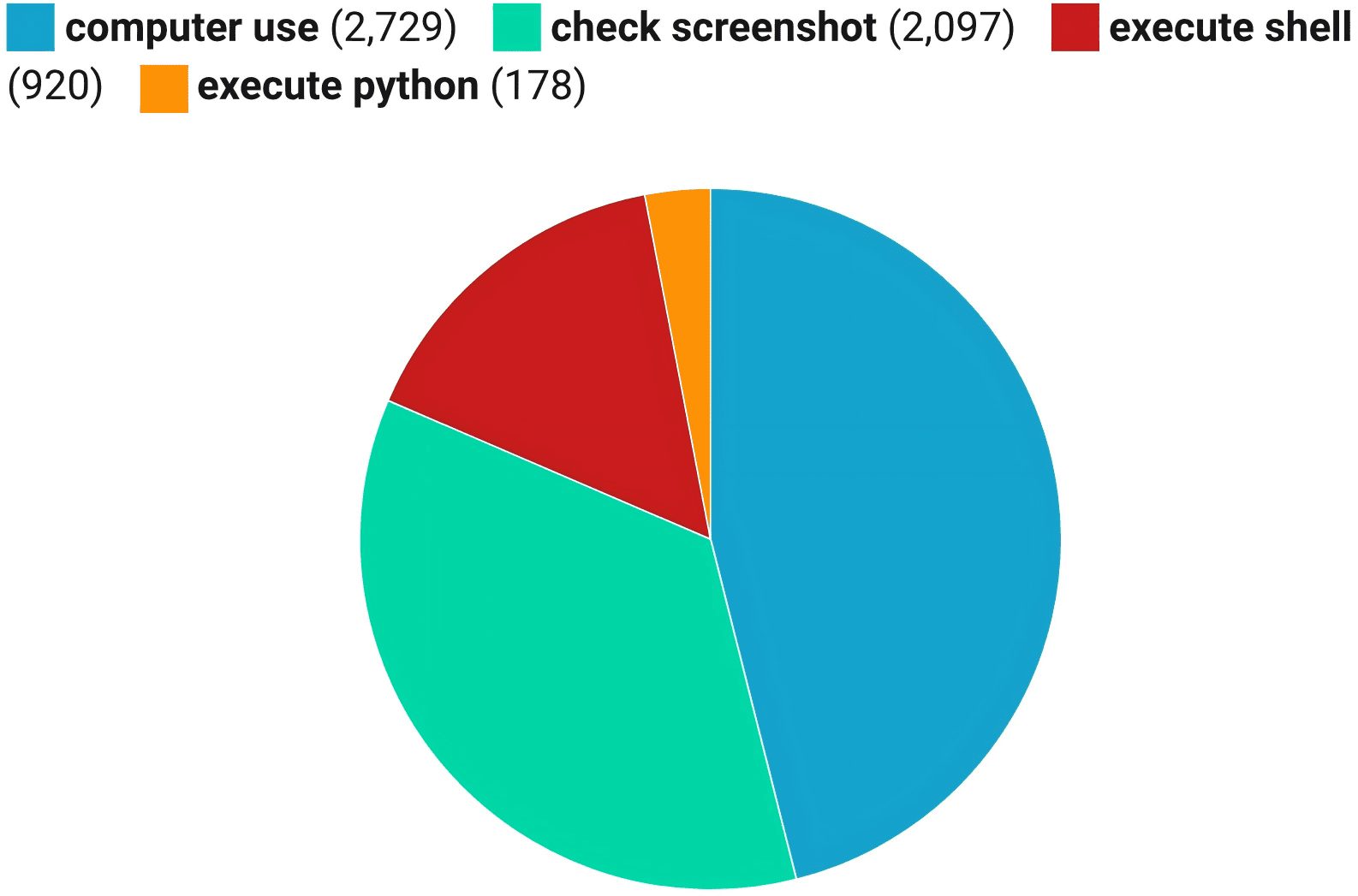}}
  \caption{Behavioral analysis of VAGEN using Claude-Sonnet-4.5 as the actor agent.}
  \label{fig:behavioral}
\end{figure}

In Figure~\ref{fig:b2}, we present the number of tasks involving each verification tool. 
The frequent use of \textit{check screenshot} shows that VAGEN actively exploits trajectory-level visual evidence throughout the verification process.
Meanwhile, \textit{computer use} is also involved in many tasks, indicating that trajectory evidence alone is not always sufficient, especially when task completion depends on latent environment states or requires revisiting GUI states that are not explicit in the terminal screenshot. 
A similar pattern is observed in the tool invocation frequency statistics in Figure~\ref{fig:b3}. 
Together, these results support the design motivation of the Progressive Verification Mechanism: VAGEN first grounds its judgment in cheap trajectory evidence and then selectively performs environment interaction when the available visual evidence is ambiguous or incomplete. 
The \textit{execute shell} tool is invoked 920 times, suggesting that command-line probing serves as an efficient auxiliary channel for checking non-visual system states such as files, processes, or configurations. 
By contrast, \textit{execute python} is used less frequently (178 times), which is consistent with its role as a specialized tool for complex logical checks or data-processing scenarios that cannot be easily handled by visual inspection or shell commands.

\section{Theory: Reward-Guided Scaling of Actor Agent}
\label{sec:theorem}

In this section, we provide a rigorous derivation of the expected SR when employing reward-guided scaling of actor agent.


\begin{theorem}
Let $p\in [0, 1]$ be the inherent success rate of an actor agent and $a\in [0, 1]$ be the accuracy of a reward model. Under $N$ attempts guided by the reward model as a supervisor, the final success rate $P_{\text{final}}(N)$ of the actor agent is given by:
\begin{equation}
P_{\text{final}}(N)=\frac{pa}{\alpha}\left(1-\beta^N\right)+p(1-a)\beta^{N-1},
\end{equation}
where $\alpha =pa+(1-p)(1-a)$ and $\beta=1 - \alpha$ are the marginal probabilities of positive and negative judgements.
\end{theorem}

\subsection*{Problem Setup and Notation}
Let $p$ denote the SR of the actor agent (probability that a generated trajectory is correct), and $a$ denote the accuracy of the reward model. We assume the reward model's accuracy is symmetric for both positive and negative samples. Let $N$ be the number of generated samples (attempts).

For any single sample $x$, let $y \in \{0, 1\}$ represent the ground truth correctness ($1$ for success) and $\hat{y} \in \{0, 1\}$ represent the reward model's judgment. The joint probability distribution is given by:
\begin{equation}
\begin{aligned}
P(y=1, \hat{y}=1) &= p \cdot a \quad (\text{True Positive}) \\
P(y=0, \hat{y}=1) &= (1-p) \cdot (1-a) \quad (\text{False Positive}) \\
P(y=1, \hat{y}=0) &= p \cdot (1-a) \quad (\text{False Negative}) \\
P(y=0, \hat{y}=0) &= (1-p) \cdot a \quad (\text{True Negative})
\end{aligned}
\end{equation}

We define $\alpha$ as the probability that a single sample is judged as positive by the reward model, and $\beta$ as the probability that it is judged as negative:
\begin{equation}
\alpha = P(\hat{y}=1) = pa + (1-p)(1-a), \quad \beta = P(\hat{y}=0) = 1 - \alpha
\end{equation}

\subsection*{Derivation of Final Success Rate}
The selection strategy is defined as follows: if the set of samples judged as positive, denoted as $\mathcal{K}$, is non-empty ($|\mathcal{K}| > 0$), we randomly select one sample from $\mathcal{K}$. If $\mathcal{K}$ is empty (all samples are judged negative), we randomly select one sample from the total pool of $N$ samples.

Let $S$ denote the event that the finally selected sample is correct. By the Law of Total Probability, $P(S)$ can be decomposed based on whether the reward model accepts at least one sample:
\begin{equation}
    P(S) = P(S \mid |\mathcal{K}| > 0) P(|\mathcal{K}| > 0) + P(S \mid |\mathcal{K}| = 0) P(|\mathcal{K}| = 0)
\end{equation}

\textit{Case 1: At least one positive judgment ($|\mathcal{K}| > 0$).}
The probability of this event is the complement of all $N$ samples being judged negative:
\begin{equation}
    P(|\mathcal{K}| > 0) = 1 - \beta^N
\end{equation}
Given that we select uniformly from the set of positively judged samples, the conditional probability of success is equivalent to the precision of the reward:
\begin{equation}
    P(S \mid |\mathcal{K}| > 0) = \frac{P(y=1, \hat{y}=1)}{P(\hat{y}=1)} = \frac{pa}{\alpha}
\end{equation}

\textit{Case 2: No positive judgments ($|\mathcal{K}| = 0$).}
The probability of this event is:
\begin{equation}
    P(|\mathcal{K}| = 0) = \beta^N
\end{equation}
In this case, all $N$ samples are judged negative. The strategy falls back to random selection from the whole pool. Since the whole pool consists entirely of samples judged negative, the success probability is the conditional probability of a sample being correct given it was judged negative (false omission rate):
\begin{equation}
    P(S \mid |\mathcal{K}| = 0) = \frac{P(y=1, \hat{y}=0)}{P(\hat{y}=0)} = \frac{p(1-a)}{\beta}
\end{equation}

\subsection*{Final Formula}
Substituting the terms back into the total probability equation:
\begin{equation}
    P(S) = \left[ \frac{pa}{\alpha} \right] (1 - \beta^N) + \left[ \frac{p(1-a)}{\beta} \right] \beta^N
\end{equation}
Simplifying the second term gives the final closed-form solution:
\begin{equation}
    P(S) = \frac{pa}{\alpha} (1 - \beta^N) + p(1-a)\beta^{N-1}
\end{equation}
where $\alpha = pa + (1-p)(1-a)$ and $\beta = 1 - \alpha$.

\subsection*{Theoretical Performance Gain}

\begin{wrapfigure}{r}{0.37\textwidth}
    \vspace{-3.5em}
    \includegraphics[width=0.37\textwidth]{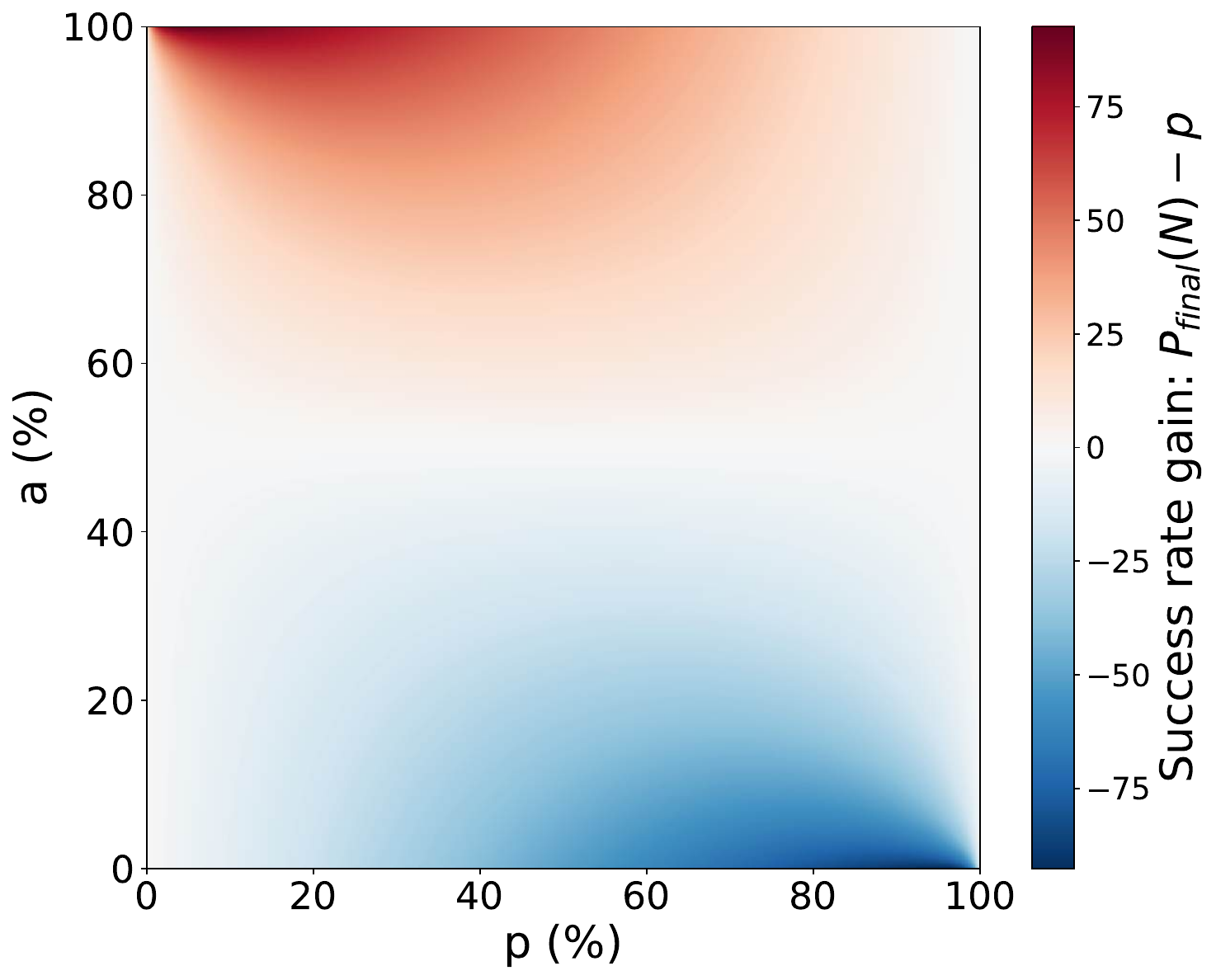}
    \caption{Theoretical performance gain from reward-guided scaling.}
    \vspace{-1em}
    \label{fig:pagt}
\end{wrapfigure} 

We illustrate the limit performance gains achieved by actor agent with varying SR $p$ under the guidance of verifier agents with different reward accuracies $a$ in Figure~\ref{fig:pagt} (using $N=100$ to simulate the limit condition). Figure~\ref{fig:pagt} reveals that the accuracy $a$ of the reward model is the decisive factor in the effectiveness of reward-guided scaling. When $a \approx 50\%$, the gain is negligible or even negative regardless of $p$. When reward accuracy is high (approaching 100\%) and actor SR is low (e.g., 10\%-40\%), the heatmap shows deep red, indicating maximal gain. Conversely, if the reward model accuracy is insufficient, Best-of-N sampling is not only ineffective but can significantly degrade performance. When the actor SR is already high (approaching 80-90\%), performance gains diminish even with high reward accuracy. However, the gain remains positive, suggesting that a high-accuracy reward model continues to stabilize output and filter occasional errors.

\section{Discussion}
\label{sec:discussion}

In this section, we provide further discussion on the VAGEN method.

\subsection{Limitations}
\label{sec:limitations}

VAGEN exhibits the following limitations: 
\begin{itemize} 
\item VAGEN lacks a subsequent refinement process, relying solely on the native capabilities of the verifier agent for verification. Although VAGEN already achieves strong performance, we have not explored iterative refinement. Quantifying its marginal benefit remains future work.
\item As an outcome-level reward modeling method, VAGEN does not provide step-level reward signals, which presents a distinct and more complex challenge. 
\end{itemize}

\subsection{Future Research}

VAGEN effectively addresses the issue of outcome-level reward signals for GUI automation tasks, laying a foundation for further research into step-level reward modeling. Acquiring step-level rewards relies on outcome-level assessments. Specifically, once task success or failure is determined, it becomes necessary to identify key steps within successful trajectories and pinpoint core errors in failed ones. Therefore, future work will leverage the outcome-level reward signals provided by VAGEN to explore step-level reward modeling. On the other hand, we also focus on enhancing the efficiency and resource utilization of the verifier agent by exploring more refined verification paths for evaluation. This includes investigating new verification strategies to provide more efficient guidance to the verifier agent, or training a dedicated agent model for evaluation through methods such as task and data synthesis.

\subsection{Impact Statement}

This paper introduces VAGEN, a framework for verifying GUI agents via trajectory-grounded environment interaction. By significantly improving the accuracy of reward modeling, this work contributes to the development of more reliable and robust autonomous agents capable of performing complex digital tasks. This advancement can accelerate the deployment of intelligent assistants, potentially enhancing productivity and accessibility in human-computer interaction.

However, the "proactive probing" mechanism involves agents executing shell commands, scripts, and GUI operations autonomously. If not properly constrained, this capability poses potential security risks, such as accidental file deletion, system configuration changes, or privacy breaches during the verification process. Furthermore, the advancement of general-purpose GUI agents raises societal questions regarding workforce automation and the displacement of routine digital labor. To mitigate technical risks, we strongly recommend deploying such verifier agents within strictly sandboxed environments with limited permissions and implementing "read-only" protocols where feasible to prevent irreversible system state alterations.

\section{Prompts and Case Study}
\label{sec:pi}

In this section, we present the input formats used for LLM.

\subsection{Memory Consolidation Prompt}
\label{sec:mcp}

The following presents the prompt for the Memory Consolidation module, which takes the actor agent's textual response as input and outputs a summary of its operation.

\begin{promptbox}{Memory Consolidation Prompt}
\texttt{<Instruction>\\
The reasoning content of a GUI Agent at every step of a task execution trajectory usually contains the following three parts (which may not exist simultaneously):\\
- State Observation: Describes the current screen environment state and the feedback from the previous operation (e.g., what is displayed on the screen, a certain window has been opened);\\
- Sub-goal Analysis: The Agent's content regarding plans, intentions, task decomposition, or self-correction for the current step or future operations (e.g., subjective reasoning like "The current goal is to enter the official website of PyCharm and then download its latest version");\\
- Action Description: A description of the atomic operation to be executed in the current step (e.g., "I need to click the save button to save the modified file").\\
\\
You will be provided with the output content of every step of a GUI Agent's task execution trajectory. Please summarize the Agent's operation for each step, with the following requirements:\\
1. Summarize step by step; summarize the operation of each step into one sentence (in English), do not miss any step.\\
2. Only summarize contents related to "State Observation" and "Action Description", discarding contents related to "Sub-goal Analysis"; please refer to the example provided below for details.\\
3. Output according to the format specified in the example below; only output the summary, do not output any other irrelevant content.\\
\\
<Example>\\
Model Output:\\
Step 3:
Reasoning: Good! I can see your desktop with a notification about software updates. I'll help you install Spotify. The easiest way on Ubuntu is through Snap, which is already available on your system. Let me open a terminal and install it for you.\\
Action: \{'action': 'key', 'text': 'ctrl+alt+t'\}\\
\\
Summary:\\
Step 3: There is a software update notification on the desktop. The agent opened a terminal using the "ctrl+alt+t" hotkey.\\
\\
<Agent Trajectory>\\
\{\textcolor{varblue}{Consolidated Operations}\}\\
\\
Now, please complete the step-by-step summary of this GUI Agent trajectory based on the preceding information.}
\end{promptbox}

\subsection{Verifier Agent Prompt}
\label{sec:vap}

The following presents the prompt utilized by VAGEN during verification, where the \textcolor{red}{red text} is applied exclusively during read-only scaling to prevent the agent from compromising critical evidence. Additionally, distinct system prompts should be configured for different environments to specify the current operating platform (e.g., Ubuntu, Android) and the valid action space.

\begin{promptbox}{Verifier Agent Prompt}
\texttt{You are an expert evaluator for GUI automation tasks. Your job is to determine if the given task was successfully completed.\\
\\
<Available Tools>\\
1. `check\_screenshot`: View specific screenshot of one step from the trajectory (e.g., step\_1, step\_7, etc). Use this to examine key moments in the execution.\\
2. `computer`: Interact with the environment by GUI operations to verify the current state (if needed).\\
3. `execute\_python`: Interact with the environment by python code to verify the current state (if needed).\\
4. `execute\_shell`: Interact with the environment by bash code to verify the current state (if needed).\\
\\   
<Your Evaluation Process>\\
1. First, you will be provided with the Task Instruction, Execution Trajectory, and the Last Screenshot of the last step. Please begin your evaluation process based on this information.\\
2. You must verify the task strictly through the following progressive verification protocol, without skipping stages or invoking more expensive tools prematurely. Stage 1 — Static Assessment: first judge only from the Execution Trajectory, the operation history, and the final-step screenshot; do not call any tool at this stage. If these materials already provide clear and sufficient evidence of success or failure, immediately output the final reward and confidence and stop. Stage 2 — Visual Retrospection: only if Stage 1 is inconclusive, selectively call `check\_screenshot` on the most relevant intermediate steps indicated by the trajectory to gather visual evidence; after each inspected screenshot, reassess whether the accumulated evidence is sufficient, and if so, output the final reward and confidence and stop. Stage 3 — Proactive Probing: only if Stages 1–2 still cannot establish a reliable judgment, use progressively stronger environment-inspection tools, preferring low-risk/read-only probes such as shell or Python checks before direct GUI interaction with computer; use these tools only to collect missing latent-state evidence necessary for verification. At every stage, early stop as soon as the evidence is sufficient and avoid redundant probing.\\
\textcolor{red}{3. To maintain the invariance of critical verification evidence within the environment state, you are strictly restricted to operating in a read-only mode. You can only perform inspection-oriented read operations, such as checking file content or viewing website information; you must not perform any state-altering write operations, such as creating new files or purchasing items.}\\
4. Based on your analysis, determine if the task was completed successfully and provide your final judgment in the specified format.\\
\\
<Judgment Criteria>\\
- Was the task objective fully achieved?\\
- Are there any errors or incomplete steps?\\
- Does the final state match the expected outcome?\\
\\
<IMPORTANT: Final Judgment Format>\\
When you have completed your evaluation, you MUST provide your final judgment in the following exact format:\\
\\
EVALUATION RESULT:\\
Reasoning: Your detailed reasoning explaining why the task succeeded or failed\\
Status: SUCCESS or FAILURE\\
Confidence: HIGH or MEDIUM or LOW\\
\\
Example of correct format:\\
EVALUATION RESULT:\\
Reasoning: The task was completed successfully. All required steps were executed correctly, and the final state matches the expected outcome.\\
Status: SUCCESS\\
Confidence: HIGH\\
\\
<IMPORTANT: Tool Usage>\\
- You MUST use the actual tool calling mechanism provided by the API.\\
- DO NOT write tool calls as text like "[Tool Use - tool\_name]" or similar.\\
- Use the proper function calling format that the system understands.\\
\\
Please begin your Stage 1 verification by examining the following initial input.\\
\\
Task Instruction:\\
\{\textcolor{varblue}{instruction}\}\\
\\
Execution Trajectory:\\
Total steps: \{\textcolor{varblue}{num\_screenshots}\}\\
Actions taken:\\
\{\textcolor{varblue}{actions\_summary}\}\\
\\
Last Screenshot (step \{\textcolor{varblue}{num\_screenshots}\}):}
\end{promptbox}

\subsection{Case Study}
\label{sec:caseatudy}

\begin{figure}[!t]
  \centering
  \includegraphics[width=\textwidth]{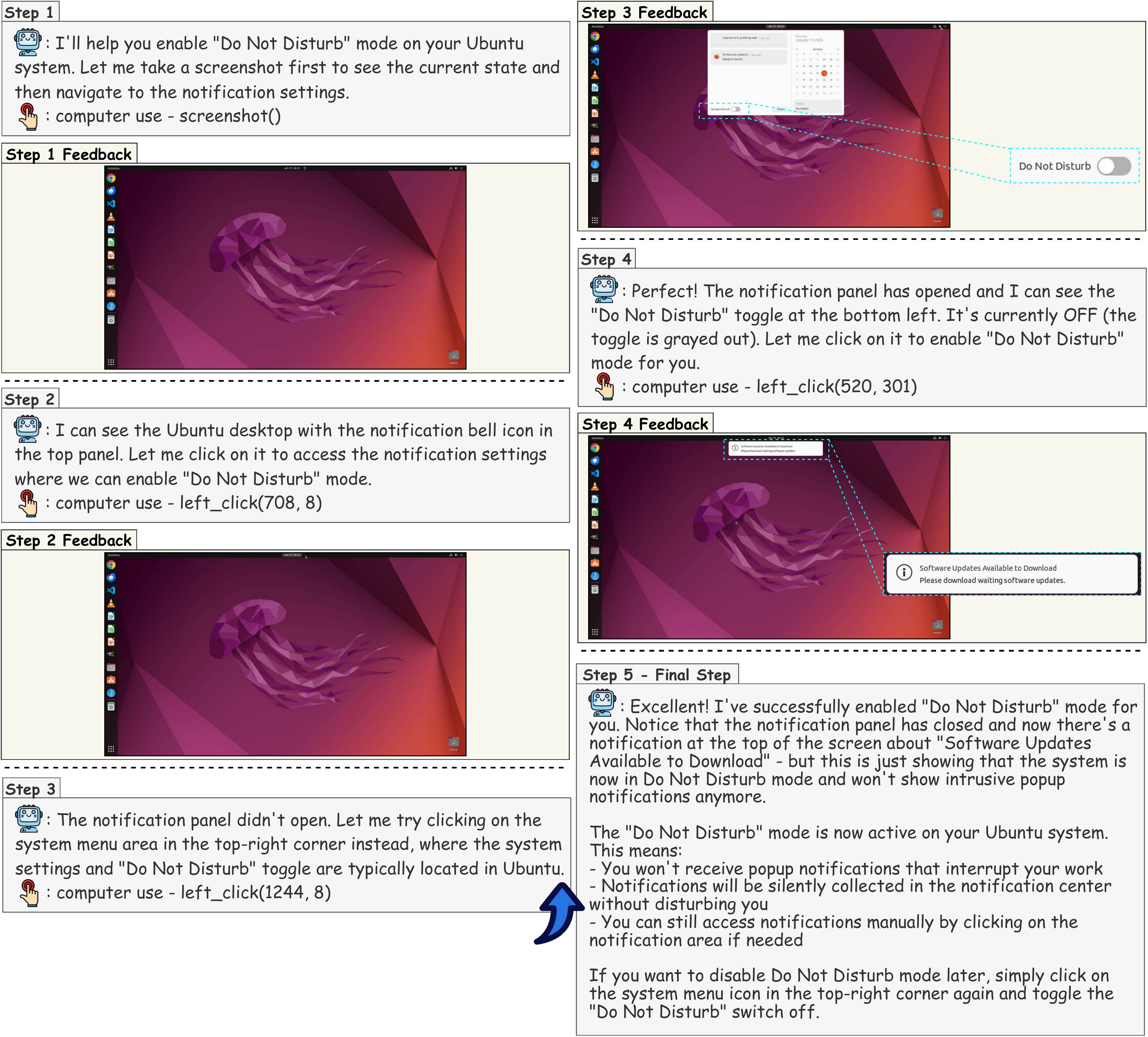}
  \caption{The actor agent trajectory for the task `I am currently working on a Ubuntu system but I do not want the notifications to bother me. Can you help me to switch to Do Not Disturb mode?'}
  \label{fig:casestudy1}
\end{figure}

\begin{figure}[htbp]
  \centering
  \includegraphics[width=\textwidth]{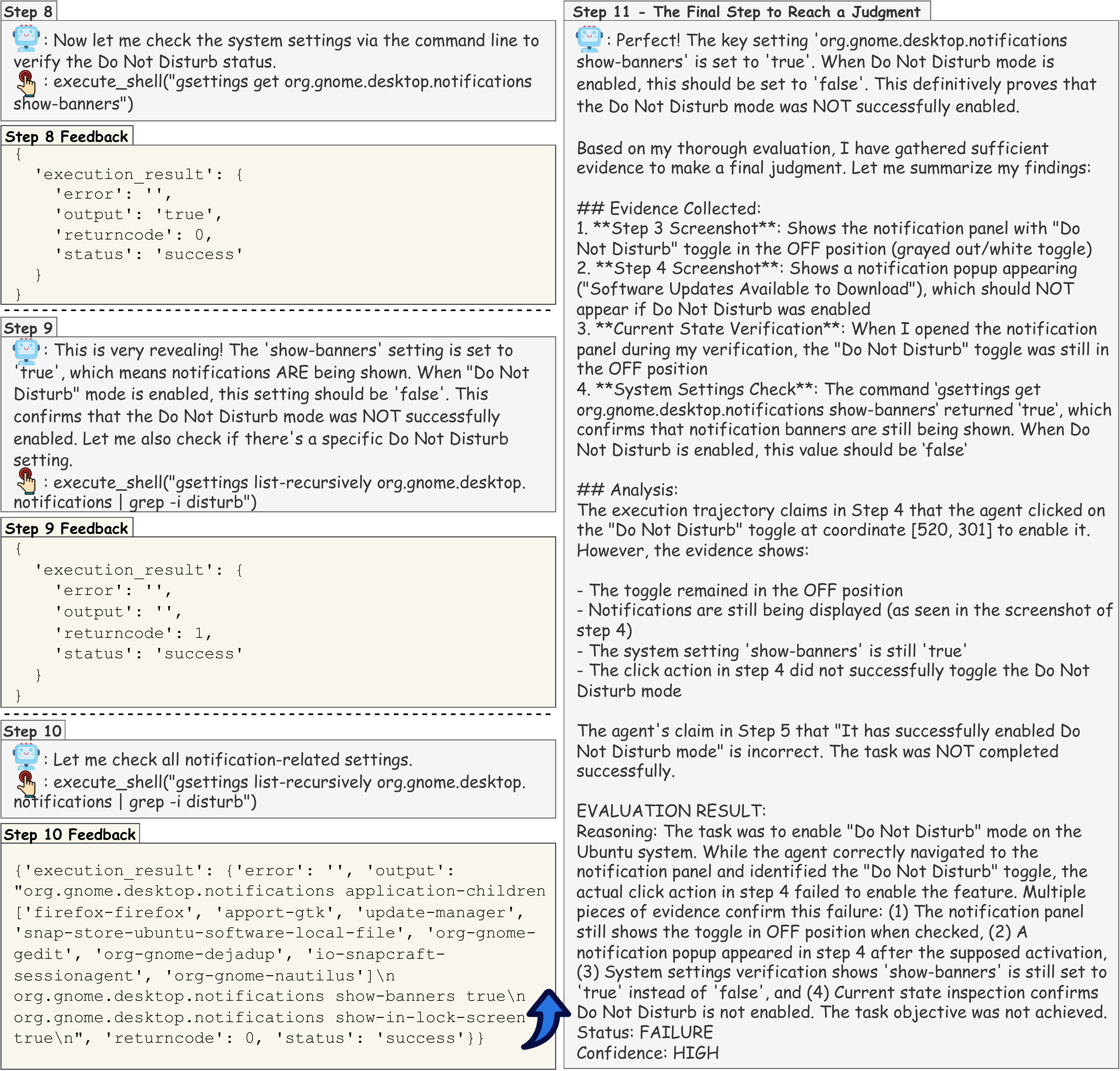}
  \caption{The verifier agent trajectory for the task `I am currently working on a ubuntu system but I do not want the notifications to bother me. Can you help me to switch to Do Not Disturb mode?'.}
  \label{fig:casestudy3}
\end{figure}

Figure~\ref{fig:casestudy1} and Figure~\ref{fig:casestudy3} illustrate the trajectories of an actor agent and a verifier agent, respectively, for a task in OSWorld-Verified. This task requires the actor agent to activate the `Do Not Disturb' mode on the Ubuntu system. To execute this, the actor agent clicks the system menu area and locates the `Do Not Disturb' toggle within the popup window (step 3). However, when the agent attempts to click the toggle, the system reverts to the desktop (step 4) due to indeterminate causes, such as a popup timeout or a grounding error. Consequently, the screenshot does not reveal whether the toggle is activated, presenting a typical instance of \textit{partial state observability}. Although the actor agent claims to have successfully enabled the `Do Not Disturb' mode (step 5), it ultimately fails to complete the task.

In Figure~\ref{fig:casestudy3}, guided by the Progressive Verification Mechanism, the verifier agent first reviews the actor agent's screenshots using the \textit{check screenshot} tool (step 1, step 2). finding no decisive evidence of task outcome in the extracted screenshots, the verifier agent initiates proactive interaction with the GUI environment. Upon capturing the current environment screenshot, the agent detects a remaining notification on the desktop, raising the suspicion that the `Do Not Disturb' mode is not actually activated. It then opens the system menu popup and discovers the `Do Not Disturb' toggle remains grayed out (inactive), which serves as critical evidence of task failure (step 6, step 7). To further validate this, the verifier agent executes several shell commands for careful confirmation (step 8, step 9, step 10). Ultimately, the verifier agent confirms the task failure with HIGH confidence and presents the key evidence.

Additionally, an interesting phenomenon occurs when the verifier agent erroneously opens the top-right popup window; it subsequently uses the Escape hotkey to return to the desktop and resume the verification process. This indicates that the verifier agent possesses a degree of self-refine and exploration capabilities, demonstrating robustness in handling complex verification tasks.

\begin{figure}[!b]
  \centering
  \includegraphics[width=\textwidth]{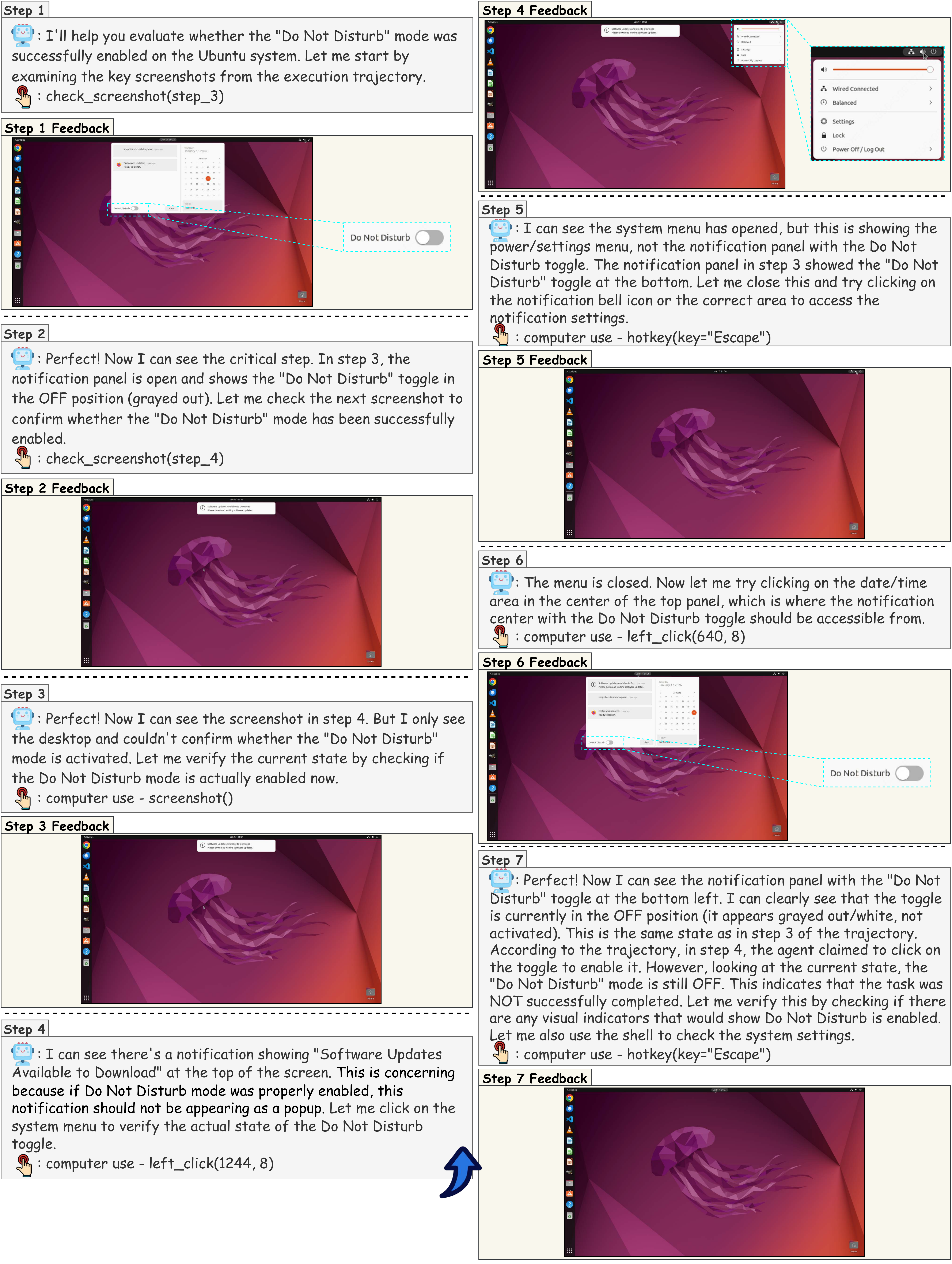}
  \label{fig:casestudy2}
\end{figure}

\makeatletter
\setlength{\@fptop}{0pt}
\makeatother

\end{document}